\documentclass[letterpaper, 10 pt, journal, twoside]{IEEEtran}

\usepackage{decar-common}
\usepackage{decar-dynamics}
\usepackage{decar-lie}
\usepackage{decar-addon}

\newsavebox{\subfigbox}

\makeatletter
\AtBeginDocument{
  \check@mathfonts
}
\makeatother

\begin{document}


\fontdimen16\textfont2=\fontdimen17\textfont2
\fontdimen13\textfont2=5pt

\title{An Adaptive Graduated Nonconvexity Loss Function for Robust Nonlinear Least-Squares Solutions}

\author{
    Kyungmin~Jung,~\IEEEmembership{Graduate~Student~Member,~IEEE,}
    Thomas~Hitchcox,~\IEEEmembership{Member,~IEEE,}
    and~James~Richard~Forbes,~\IEEEmembership{Member,~IEEE}%

    \thanks{This work was supported in part by Voyis Imaging Inc.\ through the Natural Sciences and Engineering Research Council of Canada (NSERC) Alliance program, the NSERC Discovery Grant program, and in part by the McGill Engineering Doctoral Award (MEDA) Program at McGill University.}%

    \thanks{Kyungmin~Jung, Thomas~Hitchcox, and James~Richard~Forbes are with the Department of Mechanical Engineering, McGill University, Montreal, QC, Canada, H3A 0C3.
            {\tt\footnotesize kyungmin.jung@mail.mcgill.ca, thomas.hitchcox@mail.mcgill.ca, james.richard.forbes@mcgill.ca.}}%
}


\maketitle

\begin{abstract}
    Many problems in robotics, such as estimating the state from noisy sensor data or aligning two point clouds, can be posed and solved as least-squares problems.
    Unfortunately, vanilla nonminimal solvers for least-squares problems are notoriously sensitive to outliers and initialization errors.
    The conventional approach to outlier rejection is to use a robust loss function, which is typically selected and tuned a priori.
    A newly developed approach to handle large initialization errors is graduated nonconvexity (GNC), which is defined for a particular choice of a robust loss function.
    The main contribution of this paper is to combine these two approaches by using an adaptive kernel within a GNC optimization scheme.
    This brings a solution to least-squares problems that is robust to both outliers and initialization errors, without the need for model selection and tuning.
    Simulations and experiments demonstrate that the proposed method is more robust compared to non-GNC counterparts and performs on par with other GNC-tailored loss functions.
    An example code can be found at \url{https://github.com/decargroup/gnc-adapt}.
\end{abstract}


\section{Introduction}
\label{sec:intro}

\IEEEPARstart{L}{east}-squares problems appear in robotics, computer vision, and data analytics.
However, traditional least-squares solvers perform poorly in the presence of \textit{outliers} that are often caused by spurious sensor data \cite{Roysdon2017GPS}, faulty data association \cite{Neira2001Data}, and model misspecification \cite{Cadena2016Past}.
To reduce the sensitivity to outliers during minimization of a least-squares objective function, \textit{robust loss functions} or \textit{robust kernels} \cite{DeMenezes2021Review}, such as pseudo-Huber \cite{Huber1964Robust}, Cauchy \cite{Black1996Robust}, Welsch \cite{Dennis1978Techniques}, and Geman-McClure \cite{Geman1986Baysian}, have been proposed.
However, the main drawback of these robust kernels is that they must be hand-picked and manually tuned a priori without knowledge of the actual residual distribution and are sensitive to poor initialization \cite{Hartley2004Multiple}.
Recent work by \cite{Barron2019General, Chebrolu2021Adaptive, Hitchcox2022Mind} has addressed the manual tuning issue by introducing adaptive functions that are capable of changing their shapes to fit the actual residual distribution.

Least-squares problems are also sensitive to initialization errors.
Recently, work by Yang \etal \cite{Yang2020GNC} has addressed this problem by tailoring graduated nonconvexity to truncated least-squares and Geman-McClure loss functions.
Although GNC's global optimality cannot be guaranteed, the empirical robustness of resulting robust non-minimal solvers in applications is demonstrated successfully.
However, GNC has been applied to a fixed robust loss function and the extension of GNC to generalized loss function is absent in the literature.

The main contribution of this work is a detailed derivation of the combination of an adaptive loss function and GNC (GNC-ADAPT).
Previously, GNC has been tailored to a specific loss function of choice, however, the combination of GNC with the generalized loss function no longer necessitates the specification of a particular loss function.
Such combination can be readily applied to various robotics problems such as point-cloud alignment, SLAM, pose-graph optimization, and bundle adjustment without choosing which loss function to combine with GNC.
The proposed method is tested against non-adaptive robust loss functions, adaptive loss functions, and GNC-incorporated loss functions.

The remainder of this paper is organized as follows.
Section~\ref{sec:relatedwork} presents a comprehensive review of the literature on robust loss functions and graduated nonconvexity.
Section~\ref{sec:preliminaries} provides the necessary mathematical background.
Section~\ref{sec:methodology} presents the derivation of the novel robust loss function.
The proposed methods are applied to a point-cloud alignment, a mesh registration, and a pose graph optimization problem in Section~\ref{sec:results}.
Finally, the paper is drawn to a close in Section~\ref{sec:conclusion}.


\section{Related Work}
\label{sec:relatedwork}

Robust estimation methods attempt to construct the objective function such that it is less sensitive to outliers.
The use of M-estimators modifies the nonlinear least-squares cost function with a robust loss function \cite{Barfoot2023State, Markatou1997Robust, Zhang1997Parameter, MacTavish2015At}.
This robust cost function is solved with an iteratively reweighted least-squares (IRLS) framework \cite{Chartrand2008Iteratively} that solves a sequence of weighted least-squares problems.
The convergence of the IRLS framework is guaranteed when the robust kernel meets certain criteria \cite{Aftab2015Convergence}.
For example, such a method has been proposed to reduce the sensitivity to outliers during pose graph optimization \cite{Agarwal2013Robust}.
Despite various robust kernels with particular properties \cite{Black1996Unification, Zhang1997Parameter}, the robust kernels must be hand-picked for a specific problem.

A generalized loss function that is a superset of many existing loss functions has been proposed by Barron \cite{Barron2019General} by introducing a new ``robustness'' hyperparameter.
A single continuous-valued hyperparameter in the loss function can be set such that the function is adjusted to model a wider family of problems with improved flexibility.
Moreover, this hyperparameter can be treated as an additional unknown parameter to be optimized along with the state in the generalized loss function such that manual tuning of the robust kernel is no longer required.
The adaptive loss function in \cite{Barron2019General} was originally implemented in computer vision tasks. Later, \cite{Barron2019General} was extended and applied to general nonlinear least-squares problems in \cite{Chebrolu2021Adaptive} by incorporating the hyperparameter as a part of the estimation process.
Also, Chebrolu \etal \cite{Chebrolu2021Adaptive} extended the usable range of this parameter to deal with a larger set of outlier distributions.

Recently, Hitchcox and Forbes \cite{Hitchcox2022Mind} have pointed out that most of the loss functions assume the residuals follow a Gaussian-like distribution with a mode of zero.
However, in most nonlinear least-squares problems, the residuals are defined as the norm of a multivariate error, which results in a Chi-like distribution with a nonzero mode.
Thus, applying the adaptive loss function directly to the residual distribution may lead to poor weight assignment.
This issue was addressed in \cite{Hitchcox2022Mind} by finding the mode of the Chi-like distribution and applying the adaptive loss function to the \textit{mode-shifted} residuals.
The proposed method, called the adaptive Maxwell-Boltzmann (AMB), was demonstrated in point-cloud alignment and pose averaging, and outperformed other state-of-the-art robust kernels.

Although robust kernels allow outliers to have less influence on the optimization process, they introduce local optima to the cost function and cause sensitivity to initialization, especially in the context of nonlinear models \cite{Hartley2004Multiple, Agarwal2015Robust, Grisetti2009Nonlinear}.
The sensitivity increases with asymptotically constant kernels like Geman-McClure and Welsch.
To address this problem, Yang \etal \cite{Yang2020GNC} proposed to combine nonminimal solvers with a method known as graduated nonconvexity (GNC) \cite{Black1996Unification} to solve optimization problems without requiring an accurate initial estimate.
In \cite{Yang2020GNC}, GNC and the Black-Rangarajan duality \cite{Black1996Unification} are tailored to traditional loss functions like Geman-McClure (GM) and truncated least-squares (TLS), and used with nonminimal solvers in point-cloud registration, mesh registration, pose-graph optimization, and shape alignment.
Although GNC's global optimality cannot be guaranteed, according to \cite{Yang2020GNC}, the proposed method was able to solve the aforementioned problems without requiring an accurate state prior to optimization, outperforming techniques like RANSAC \cite{Fischler1981RANSAC}.
The use of GNC with robust kernels has been further explored in McGann's work \cite{McGann2023Robust} by proposing the Scale Invariant Graduated (SIG) kernel, which is a modification of the Geman-McClure loss function.
Further, Choi \etal \cite{Choi2023Adaptive} extends GNC-SIG to uniformly increase the nonconvexity using B-splines.
However, this work is still limited to a combination of GNC with a specific loss function, Geman-McClure.
The extension of the GNC method to generalized loss function is therefore absent in the literature.

\section{Preliminaries}
\label{sec:preliminaries}

This section introduces the mathematical preliminaries of a least-squares problem, a robust loss function, an adaptive kernel, and graduated nonconvexity.

\subsection{Least-squares Problem}
\label{sec:preliminaries_leastsquares}

Let ${\mbf{e}_{i}=\mbf{e}(\mbf{y}_{i}, \mbf{g}(\mbf{x}))}$ be an error function that quantifies the difference between the ${i}$-th measurement ${\mbf{y}_{i}}$ and the expected measurement given a measurement model ${\mbf{g}(\cdot)}$ and a state ${\mbf{x}}$.
The Mahalanobis distance of the error function is 
\begin{align}
    \epsilon_{i}\left(\mbf{y}_{i}, \mbf{x}\right) & = \norm{\mbf{e}_{i}}_{\mbs{\Sigma}_{i}^{-1}} = \sqrt{\mbf{e}_{i}^{\trans}\mbs{\Sigma}_{i}^{-1}\mbf{e}_{i}},
    \label{eq:mahalanobis}
\end{align}
where ${\mbs{\Sigma}_{i}}$ is the covariance on the ${i}$-th error.
For brevity of notation, ${\epsilon_{i} = \epsilon_{i}\left(\mbf{y}_{i}, \mbf{x}\right)}$.

\subsection{Robust Loss Function}
\label{sec:preliminaries_robustlossfunctions}

Consider the following least-squares problem,
\begin{align}
    \mbf{x}^{\star} & = \underset{\mbf{x}\in\mc{X}}{\argmin} \sum_{i=1}^{N} \onehalf \epsilon_{i}^{2},
    \label{eq:ls}
\end{align}
where ${\mbf{x} \in \mc{X}}$ is the state to be estimated and ${N}$ is the number of measurements.
In the presence of outliers, \eqref{eq:ls} provides a poor estimate of ${\mbf{x}}$.
Robust least-squares solvers substitute the quadratic cost in \eqref{eq:ls} with a loss function ${\rho(\cdot)}$,
\begin{align}
    \mbf{x}^{\star} & = \underset{\mbf{x}\in\mc{X}}{\argmin} \sum_{i=1}^{N} \rho\left( \epsilon_{i} \right).
    \label{eq:loss_function}
\end{align}
Equation~\ref{eq:loss_function} can be solved using an iteratively reweighted least-squares (IRLS) framework \cite{Chartrand2008Iteratively} that solves a sequence of weighted least-squares problems,
\begin{align}
    \mbf{x}^{\star} & = \underset{\mbf{x}\in\mc{X}}{\argmin} \sum_{i=1}^{N} \onehalf w_{i} \epsilon_{i}^{2}.
    \label{eq:IRLS}
\end{align}
The weights ${w_{i}}$ in \eqref{eq:IRLS} are obtained by equating the gradients of the loss function in \eqref{eq:loss_function} and \eqref{eq:IRLS} with respect to ${\mbf{x}}$,

\begin{subequations}
    \begin{align}
        \pd{\rho\left(\epsilon_{i}\right)}{\epsilon_{i}} \pd{\epsilon_{i}}{\mbf{x}} & = w_{i} \epsilon_{i} \pd{\epsilon_{i}}{\mbf{x}},                       \\
        w_{i}                                                                       & = \f{1}{\epsilon_{i}}\pd{\rho\left(\epsilon_{i}\right)}{\epsilon_{i}}.
        \label{eq:weights}
    \end{align}
\end{subequations}
This approach allows standard solvers like Gauss-Newton and Levenberg-Marquardt algorithms to solve \eqref{eq:loss_function}.

\subsection{Adaptive Robust Loss}
\label{sec:preliminaries_adaptiveloss}

This section discusses the formulation of the adaptive loss function proposed by Barron in \cite{Barron2019General}, its limitations, and the solutions to the limitations proposed by Chebrolu \etal in \cite{Chebrolu2021Adaptive}.
The simplest form of the adaptive loss function is
\begin{align}
    \rho\left(\epsilon, \alpha\right) & = \f{\abs{\alpha - 2}}{\alpha} \left( \left(\f{\epsilon^2}{\abs{\alpha-2}}+1\right)^{\nicefrac{\alpha}{2}} - 1 \right),
    \label{eq:barron_adaptive_loss}
\end{align}
where ${\alpha\in(-\infty,2]}$ is a shape parameter that controls the robustness of the loss.
Accounting for singularities, \eqref{eq:barron_adaptive_loss} is rewritten in a piecewise function as
\begin{align}
    \rho\left(\epsilon, \alpha\right) & =
    \begin{cases}
        \onehalf\epsilon^{2}                                                                                                & \textrm{if } \alpha = 2,       \\
        \log\left(\onehalf\epsilon^{2} + 1\right)                                                                           & \textrm{if } \alpha = 0,       \\
        1 - \exp\left(-\onehalf\epsilon^{2}\right)                                                                          & \textrm{if } \alpha = -\infty, \\
        \f{\abs{\alpha-2}}{\alpha}\left(\left(\f{\epsilon^{2}}{\abs{\alpha-2}} + 1\right)^{\nicefrac{\alpha}{2}} - 1\right) & \textrm{otherwise}.
    \end{cases}
    \label{eq:barron_adaptive_loss_piecewise}
\end{align}
In Barron's work, ${\alpha}$ is treated as an additional unknown parameter to be optimized along with ${\mbf{x}}$ in the generalized loss function,
\begin{align}
    \mbf{x}^{\star}, \alpha^{\star} & = \underset{\mbf{x}\in\mc{X}, \alpha\in(-\infty,2]}{\argmin} \sum_{i=1}^{N} \rho\left(\epsilon_{i}, \alpha\right).
    \label{eq:barron_adaptive_loss_solution}
\end{align}
However, naively solving for \eqref{eq:barron_adaptive_loss_solution} can result in a trivial solution with some ${{\alpha}^{\star}}$ that downweights all the residuals without affecting ${\mbf{x}}$.
Thus, \cite{Barron2019General} avoids this by adding a regularization term derived from the probability distribution,
\begin{subequations}
    \begin{align}
        \label{eq:barron_adaptive_loss_probability}
        p\left(\epsilon \mid \bar{\epsilon}, \alpha\right) & = \f{1}{Z\left(\alpha\right)} \exp\left(-\rho\left(\epsilon - \bar{\epsilon}, \alpha\right)\right),         \\
        \label{eq:barron_partition}
        Z\left(\alpha\right)                            & = \int_{-\infty}^{\infty} \exp\left(-\rho\left(\epsilon \mid \bar{\epsilon}, \alpha\right)\right)\dee\epsilon,
    \end{align}
\end{subequations}
where ${\bar{\epsilon}}$ is the residual mean and ${Z(\alpha)}$ is a partition function.
The general adaptive loss with the regularization term is constructed from the negative log-likelihood of \eqref{eq:barron_adaptive_loss_probability},
\begin{subequations}
    \begin{align}
        \rho_{a}\left(\epsilon, \alpha\right) & = -\log\left( p\left(\epsilon \mid \bar{\epsilon}, \alpha\right) \right)          \\
                                              & = \rho\left(\epsilon, \alpha\right) + \log\left( Z\left(\alpha\right) \right).
        \label{eq:barron_general_adaptive_loss}
    \end{align}
\end{subequations}
However, \eqref{eq:barron_general_adaptive_loss} is only defined for ${\alpha \geq 0}$ because ${Z(\alpha)}$ diverges when ${\alpha < 0}$.

To solve this issue associated with $\alpha < 0$, \cite{Chebrolu2021Adaptive} computes an approximate partition function ${\tilde{Z}(\alpha)}$ by limiting the integral bounds with a hyperparameter ${\tau}$ such that
\begin{align}
    \tilde{Z}\left(\alpha\right) & = \int_{-\tau}^{\tau} \exp\left( -\rho\left(\epsilon \mid \bar{\epsilon}, \alpha\right) \right) \dee\epsilon.
    \label{eq:cheb_normalization}
\end{align}
This allows the shape parameter to be dynamically adapted for values between ${-\infty}$ and ${2}$.
The truncation bound is often set for a specific problem.
For example, if the magnitude of the residuals is expected to be large, a larger ${\tau}$ value is used.
Further, the work by Chebrolu \etal \cite{Chebrolu2021Adaptive} finds the optimal shape parameter through a grid search by solving the optimization problem
\begin{subequations}
    \begin{align}
        \alpha^{\star} & = \underset{\alpha\in(-\infty,2]}{\argmin} -\log\left( p\left(\epsilon \mid \bar{\epsilon}, \alpha\right) \right)                                         \\
                       & = \underset{\alpha\in(-\infty,2]}{\argmin} N \log\left( \tilde{Z}\left(\alpha\right) \right) + \sum_{i=1}^{N} \rho\left( \epsilon_{i}, \alpha\right).
        \label{eq:cheb_alphastar}
    \end{align}
\end{subequations}
Having ${{\alpha}^{\star}}$ and the loss function \eqref{eq:barron_general_adaptive_loss}, the weights ${w_{i}}$ are obtained using \eqref{eq:weights},
\begin{align}
    w\left(\epsilon, \alpha^{\star}\right) & =
    \begin{cases}
        1                                                                                              & \textrm{if } \alpha^{\star} = 2,       \\
        \f{2}{\epsilon^{2} + 2}                                                                        & \textrm{if } \alpha^{\star} = 0,       \\
        \exp\left(-\onehalf\epsilon^{2}\right)                                                         & \textrm{if } \alpha^{\star} = -\infty, \\
        \left( \f{\epsilon^{2}}{\abs{\alpha^{\star}-2}} + 1 \right)^{\nicefrac{\alpha^{\star}}{2} - 1} & \textrm{otherwise}.
    \end{cases}
    \label{eq:arlf_weights}
\end{align}

\subsection{Adaptive Maxwell-Boltzmann Loss}
\label{sec:preliminaries_amb}

Provided the error is normally distributed, the residuals will follow a Chi distribution with some nonzero mode value ${\tilde{\epsilon}}$.
However, most of the existing robust loss functions are derived based on Gaussian-like distribution with its peak at zero.
Therefore, the residuals will be weighted the highest near zero and weighted less as they get further away from zero.
Recently, Hitchcox and Forbes \cite{Hitchcox2022Mind} discovered that this Gaussian assumption results in lower weights assigned to residuals clustered around the non-zero mode value.
As such, \cite{Hitchcox2022Mind} proposed the AMB loss function that addresses this problem by first fitting an ${n_{e}}$-dimensional Maxwell-Boltzmann (MB) distribution to the residuals,
\begin{align}
    p_{\textrm{MB}}\left(\epsilon \mid a, n_{e}\right) & = \f{\epsilon^{n_{e}-1}}{a^{n_{e}} 2^{\left(\nicefrac{n_{e}}{2} - 1\right)} \Gamma\left(\nicefrac{n_{e}}{2}\right)} \exp\left( -\nicefrac{\epsilon^{2}}{2a^{2}} \right),
    \label{eq:pmb}
\end{align}
where ${n_{e}}$ is the dimension of the residual and ${a}$ is the MB shape parameter.
The optimal shape parameter ${a^{\star}}$ can be found by minimizing the negative log-likelihood of \eqref{eq:pmb} with respect to ${a}$.
However, this could result in poor estimation in the presence of outliers.
Thus, ${a^{\star}}$ is found by solving
\begin{align}
    a^{\star} & = \underset{a\in\mathbb{R}^{+}}{\argmin} \int_{0}^{\tau} \left( q\left(\epsilon\right) \cdot \left(p_{\textrm{MB}}\left(\epsilon \mid a, n_{e}\right) - q\left(\epsilon\right)\right) \right)^{2} \dee\epsilon,
    \label{eq:a_star}
\end{align}
where ${q(\epsilon)}$ is the residual distribution.
This results in a better fit in high-frequency inlier areas.
Equation~\eqref{eq:a_star} is solved analytically using Newton's method with a line search \cite{Hitchcox2022Mind}.
The optimal shape parameter is used to find the nonzero mode value,
\begin{align}
    \tilde{\epsilon} & = a^{\star} \sqrt{n_{e} - 1}.
    \label{eq:mb_mode}
\end{align}
Residuals satisfying ${\epsilon \leq \tilde{\epsilon}}$ are considered inliers and assigned a weight equal to ${1}$, and any residuals satisfying ${\epsilon > \tilde{\epsilon}}$ are assigned a weight that is strictly less than one in an adaptive fashion.
To do so, the optimal shape parameter ${\alpha^{\star}}$ that fits the \textit{mode-shifted} residuals, ${\xi = \epsilon - \tilde{\epsilon}}$, is computed using \eqref{eq:cheb_alphastar}
\begin{align}
    \alpha^{\star} = \underset{\alpha\in(-\infty,2]}{\argmin} M \log\left( \tilde{Z}\left(\alpha\right) \right) + \sum_{i=1}^{M} \rho\left(\xi_{i}, \alpha\right),
    \label{eq:amb_alpha}
\end{align}
where ${M}$ is the total number of ${\xi_{i}}$.
The partition function ${\tilde{Z}(\cdot)}$ is now computed with the \textit{mode-shifted truncation bound}, ${\tilde{\tau} = \tau - \tilde{\epsilon}}$,
\begin{align}
    \tilde{Z}\left(\alpha\right) = \int_{0}^{\tilde{\tau}} \exp\left( -\rho\left(\xi \mid \mu_{\xi}, \alpha\right) \right) \dee\xi.
    \label{eq:amb_partition}
\end{align}
Thus, given ${\tilde{\epsilon}}$ and ${\alpha^{\star}}$, the weights are
\begin{align}
    \tilde{w}\left(\epsilon, \tilde{\epsilon}, \alpha^{\star}\right) & =
    \begin{cases}
        1                                                         & \textrm{if } \epsilon \leq \tilde{\epsilon}, \\
        w\left(\epsilon - \tilde{\epsilon}, \alpha^{\star}\right) & \textrm{if } \epsilon > \tilde{\epsilon}.
    \end{cases}
    \label{eq:amb_weights}
\end{align}
These weights ${w(\cdot)}$ in \eqref{eq:amb_weights} are from \eqref{eq:arlf_weights}.

\subsection{Black-Rangarajan Duality and Graduated Nonconvexity}
\label{sec:preliminaries_brdgnc}

This section revisits the Black-Rangarajan duality between robust estimation and outlier process, and the application of graduated nonconvexity to nonconvex optimization problems.
Black-Rangarajan duality has been used to solve least-squares problems with robust kernels.
\begin{lemma}[Black-Rangarajan Duality \cite{Black1996Unification}]
    \label{lem:brd}
    Given a loss function ${\rho(\cdot)}$, define some function ${\phi(z)\coloneqq\rho(\sqrt{z})}$. If ${\phi(z)}$ satisfies
    \begin{subequations}
        \begin{align}
            \label{eq:brd_conditions}
            \lim_{z\to{0}} \phi'(z)      & = \onehalf, \\
            \lim_{z\to{\infty}} \phi'(z) & = 0,        \\
            \phi''(z)                    & < 0,
        \end{align}
    \end{subequations}
    then there exists an analytical outlier process function ${\Phi_{\rho}(\cdot)}$ that can be written as
    \begin{align}
        \rho\left(\epsilon\right) & = \onehalf w \epsilon^2 + \Phi_{\rho}\left(w\right).
    \end{align}
\end{lemma}

Thus, \eqref{eq:IRLS} can be rewritten as
\begin{align}
    \mbf{x}^{\star}, \mbf{w}^{\star} & = \underset{\mbf{x}\in\mc{X}, w_{i}\in(0, 1]}{\argmin} \sum_{i=1}^{N} \left( \onehalf w_{i} \epsilon_{i}^{2} + \Phi_{\rho}\left(w_{i}\right)\right),
    \label{eq:brd}
\end{align}
where ${w_{i}}$ is the weight associated to ${i}$-th measurement, ${\mbf{w}^{\star}}$ is the vector form of optimal weights,
and the outlier process function ${\Phi_{\rho}(\cdot)}$ defines a penalty on ${w_{i}}$ \cite{Black1996Unification}.
An appropriate regularization function ${\Phi_{\rho}(\cdot)}$ must be found for a particular loss function ${\rho(\cdot)}$.
The conditions on ${\rho(\cdot)}$ in \eqref{eq:brd_conditions} are satisfied by many common robust loss functions \cite{Black1996Unification}.

This duality converts \eqref{eq:loss_function} into polynomials by regularizing \eqref{eq:IRLS} with an additional constraint ${\Phi_{\rho}(\cdot)}$, and also provides an analytical expression for ${\Phi_{\rho}(\cdot)}$.
However, the duality does not enable solving a highly nonconvex optimization problem with poor initial estimates.
This nonconvexity can be tackled by using GNC \cite{Blake1987Visual}.

The idea behind GNC is to choose a surrogate function ${\rho_{\mu}(\cdot)}$ with a control parameter ${\mu}$ that changes the shape of ${\rho(\cdot)}$ such that \eqref{eq:brd} can be rewritten as
\begin{align}
    \mbf{x}^{\star}, \mbf{w}^{\star} & = \underset{\mbf{x}\in\mc{X}, w_{i}\in(0,1]}{\argmin} \sum_{i=1}^{N} \left(\onehalf w_{i} \epsilon_{i}^{2} + \Phi_{\rho_{\mu}}\left(w_{i}\right)\right),
    \label{eq:brd_gnc}
\end{align}
where ${\Phi_{\rho_{\mu}}(\cdot)}$ is the outlier process function for the surrogate loss function, ${\rho_{\mu}(\cdot)}$.
The parameter ${\mu}$ allows a convex approximation of ${\rho(\cdot)}$ to be obtained such that it can be readily minimized.
GNC solves the nonconvex problem by gradually increasing the nonconvexity of ${\rho_{\mu}(\cdot)}$ during optimization until it recovers the original function.
In Yang \etal's work \cite{Yang2020GNC}, \eqref{eq:brd_gnc} is solved using IRLS by modifying ${\mu}$ at every iteration.

\section{Methodology}
\label{sec:methodology}

Although applying graduated nonconvexity theory to various loss functions has improved the robustness to poor initialization, the choice of loss function must still be selected by the user, which may lack motivation until data is collected and analyzed.
This section presents the derivation of the proposed adaptive graduated nonconvexity (GNC-ADAPT) loss function.
The proposed loss function no longer requires the user to hand-pick a specific loss function for different problems, while also providing robustness to poor initialization via incorporating GNC.

Assuming that the optimal shape parameter ${\alpha^{\star}}$ is precomputed using \eqref{eq:cheb_alphastar}, GNC is incorporated into \eqref{eq:barron_adaptive_loss}.
Based on the properties inherited from GNC, optimization with the proposed loss function is more likely to converge to a global minimum than optimization using non-GNC counterparts.

\subsection{GNC Surrogate Function to Adaptive Robust Loss}
\label{sec:methodology_surrogate}

This section presents the main contribution of the paper, the combination of GNC with the adaptive loss function.
Suppose there exists a surrogate function of \eqref{eq:barron_adaptive_loss} of the form
\begin{align}
    \rho_{\mu}\left(\epsilon, \alpha^{\star}\right) & = \f{\abs{f-2}}{f}\left(\left(\f{\epsilon^{2}}{\abs{f-2}} + 1\right)^{\nicefrac{f}{2}} - 1\right),
    \label{eq:rlf_surrogate}
\end{align}
where ${f=f(\mu, \alpha^{\star})}$ is a \textit{shape function} that takes in the optimal shape parameter ${\alpha^{\star}}$ and the convexity control parameter ${\mu}$ as inputs to determine the shape of the adaptive loss function.
Recall that GNC computes a solution to the nonconvex problem by starting from its convex surrogate and gradually increasing its nonconvexity until the original function is recovered.
Notice that ${\rho_{\mu}(\cdot)}$ from \eqref{eq:rlf_surrogate} becomes quadratic in the limit of ${f(\cdot) \to 2}$, and that the original ${\rho(\cdot)}$ from \eqref{eq:barron_adaptive_loss} is recovered in the limit of ${f(\cdot) \to \alpha^{\star}}$.
Also, ${f(\cdot)\leq{2}}$ for all ${\mu}$ because \eqref{eq:barron_adaptive_loss} is only defined for ${\alpha\in(-\infty,2]}$.
As such, the shape function ${f(\cdot)}$ must satisfy the conditions,
\begin{subequations}
    \label{eq:agnc_condition}
    \begin{align}
        \label{eq:agnc_condition_1}
        \lim_{\mu\to{c_{1}}} f\left(\mu, \alpha^{\star}\right) & = 2,              \\
        \label{eq:agnc_condition_2}
        \lim_{\mu\to{c_{2}}} f\left(\mu, \alpha^{\star}\right) & = \alpha^{\star}, \\
        \label{eq:agnc_condition_3}
        f\left(\mu, \alpha^{\star}\right) \leq 2.
    \end{align}
\end{subequations}
The constants ${c_{1}}$ and ${c_{2}}$ in \eqref{eq:agnc_condition_1} and \eqref{eq:agnc_condition_2} are user-defined and determine the shape of ${f(\cdot)}$.
The performance of the adaptive GNC loss function depends on the shape of ${f(\cdot)}$ as the shape function governs the rate of nonconvexity added to the function, which in turn affects the convergence rate of the optimization process.
The shape functions allow the algorithm to adapt to different outlier distributions and provide a robust solution to the optimization problem.
In the following examples, three different shape functions and their properties are discussed.
The purpose of these examples is to show how the shape functions can be formulated based on the conditions specified in \eqref{eq:agnc_condition}, but not to perform an exhaustive analysis on which shape function performs the best.

\begin{example}
    \label{ex:1}
    Let ${c_{1}=\infty}$ and ${c_{2}=1}$ such that the original function is restored as ${\mu}$ decreases from ${\infty}$ to ${1}$.
    Amongst many, a shape function ${f(\cdot)}$ can be formulated as
    \begin{align}
        f\left(\mu, \alpha^{\star}\right) & = \f{\alpha^{\star} + 2\mu - 2}{\mu},
        \label{eq:agnc_sf1}
    \end{align}
    such that
    \begin{subequations}
        \begin{align}
            \label{eq:agnc_ex1_conditions_1}
            \lim_{\mu\to{\infty}} f\left(\mu, \alpha^{\star}\right) & = \lim_{\mu\to{\infty}} \f{\alpha^{\star} + 2\mu - 2}{\mu} = 2,         \\
            \label{eq:agnc_ex1_conditions_2}
            \lim_{\mu\to{1}} f\left(\mu, \alpha^{\star}\right)      & = \lim_{\mu\to{1}} \f{\alpha^{\star} + 2\mu - 2}{\mu} = \alpha^{\star}, \\
            \label{eq:agnc_ex1_conditions_3}
            f\left(\mu, \alpha^{\star}\right)                       & \leq 2, \ \forall \mu \in [1, \infty).
        \end{align}
    \end{subequations}
    However, when ${\alpha^{\star}=-\infty}$, \eqref{eq:agnc_ex1_conditions_1} and \eqref{eq:agnc_ex1_conditions_2} are undefined.
    Thus, \eqref{eq:agnc_sf1} is valid for ${\alpha^{\star}\in(-\infty,2]}$.
    In practice, the convexity control parameter ${\mu}$ decreases from some initial value ${\mu_{0} = \max\left(\epsilon^{2}\right)}$ to ${1}$ by
    \begin{align}
        \label{eq:continuation}
        \mu_{k} = \left(\mu_{k-1} - 1\right) / c + 1,
    \end{align}
    where ${c > 1}$ is some constant.
\end{example}
\begin{example}
    \label{ex:2}
    Alternatively, let ${c_{1}=0}$ and ${c_{2}=\infty}$ to represent increasing ${\mu}$.
    Then, a shape function ${f(\cdot)}$ can be formulated as
    \begin{align}
        f\left(\mu, \alpha^{\star}\right) & = \alpha^{\star} \exp\left( \f{1}{-\mu} \right) + 2 \exp\left(-\mu\right),
        \label{eq:agnc_sf2}
    \end{align}
    such that
    \begin{subequations}
        \begin{align}
            \label{eq:agnc_ex2_conditions_1}
            \lim_{\mu\to{0^{+}}} f\left(\mu, \alpha^{\star}\right)  & = \lim_{\mu\to{0^{+}}} \alpha^{\star} \exp\left( \f{1}{-\mu} \right) + 2 \exp\left(-\mu\right) = 2,               \\
            \label{eq:agnc_ex2_conditions_2}
            \lim_{\mu\to{\infty}} f\left(\mu, \alpha^{\star}\right) & = \lim_{\mu\to{\infty}} \alpha^{\star} \exp\left( \f{1}{-\mu} \right) + 2 \exp\left(-\mu\right) = \alpha^{\star}, \\
            \label{eq:agnc_ex2_conditions_3}
            f\left(\mu, \alpha^{\star}\right)                       & \leq 2, \ \forall \mu \in \mathbb{R}^{+}.
        \end{align}
    \end{subequations}
    Again, for ${\alpha^{\star}=-\infty}$, \eqref{eq:agnc_ex2_conditions_1} and \eqref{eq:agnc_ex2_conditions_1} are not satisfied.
    With this shape function, the convexity parameter ${\mu}$ increases from some initial value ${\mu_{0} = \f{1}{\max\left(\epsilon^{2}\right)}}$ to ${\infty}$ by
    \begin{align}
        \mu_{k} = c \mu_{k-1},
    \end{align}
    where ${c > 1}$ is some constant.
\end{example}
\begin{example}
    \label{ex:3}
    With the same choice of ${c_{1}}$ and ${c_{2}}$ as in Example~\ref{ex:2}, a different shape function ${f(\cdot)}$ can be formulated as
    \begin{align}
        f\left(\mu, \alpha^{\star}\right) & = \f{\alpha^{\star} \mu + 2}{\mu + 1},
        \label{eq:agnc_sf3}
    \end{align}
    such that
    \begin{align}
        \lim_{\mu\to{0}^{+}} f\left(\mu, \alpha^{\star}\right) & = \lim_{\mu\to{0}^{+}} \f{\alpha^{\star} \mu + 2}{\mu + 1} = 2,               \\
        \lim_{\mu\to\infty} f\left(\mu, \alpha^{\star}\right)  & = \lim_{\mu\to{\infty}} \f{\alpha^{\star} \mu + 2}{\mu + 1} = \alpha^{\star}, \\
        f\left(\mu, \alpha^{\star}\right)                      & \leq 2, \ \forall \mu \in \mathbb{R}^{+}.
    \end{align}
\end{example}

\begin{figure}[tb]
    \centering
    \includegraphics[width=\linewidth,clip=true,trim={0cm 0cm 0cm 0cm}]{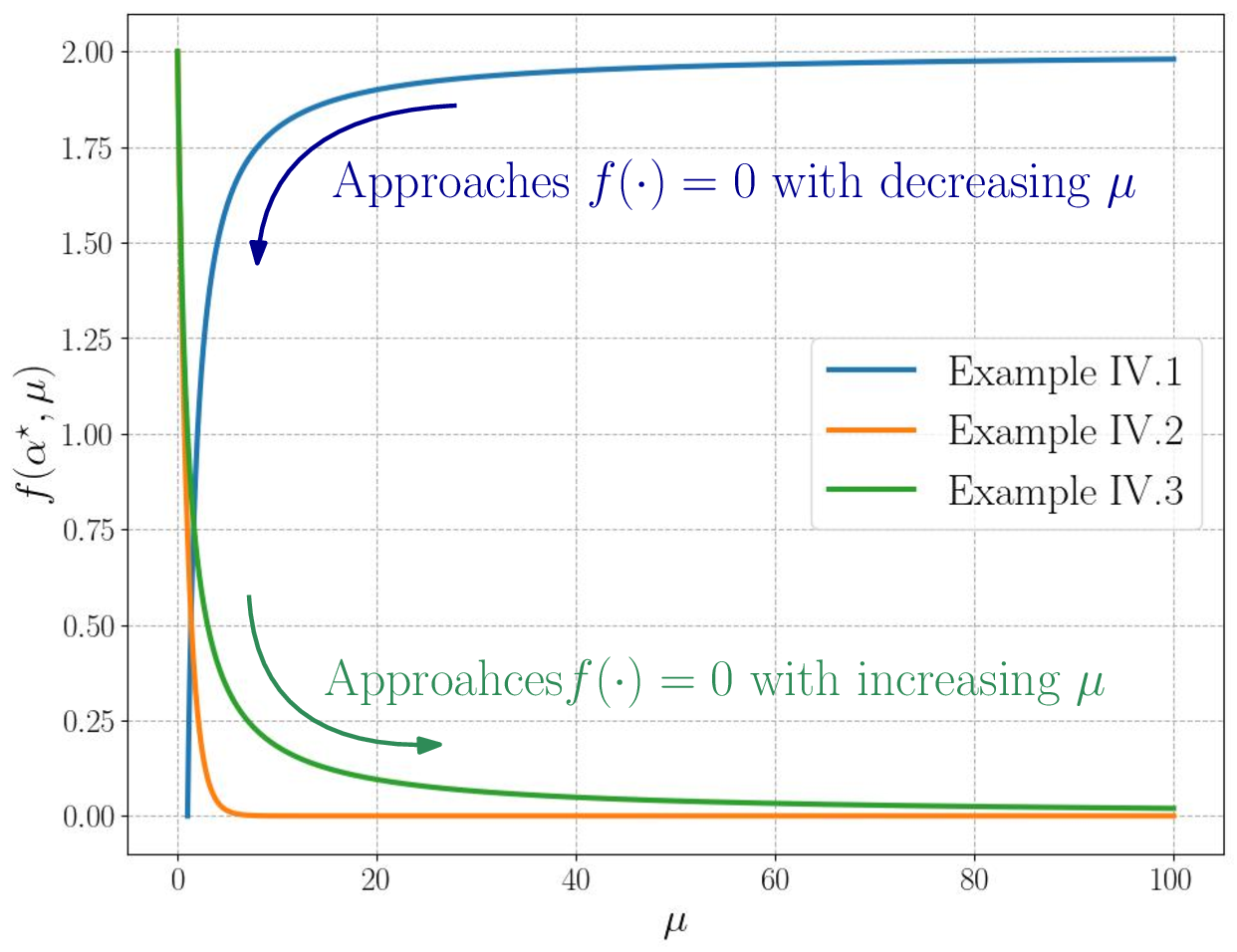}
    \caption{
        Shape functions with ${\alpha^{\star}=0}$ for Examples~\ref{ex:1}, \ref{ex:2}, and \ref{ex:3} shown in blue, orange, and green lines, respectively.
        Blue line approaches ${f(\cdot)=\alpha^{\star}}$ as ${\mu}$ decreases towards ${1}$, whereas orange and green lines approach ${f(\cdot)=\alpha^{\star}}$ as ${\mu}$ increases towards ${\infty}$.
        However, the rate at which ${f(\cdot)}$ approaches ${\alpha^{\star}}$, also interpreted as the amount of nonconvexity added to the surrogate loss function ${\rho_{\mu}(\cdot)}$, is higher in the orange line (Example~\ref{ex:2}).
    }
    \label{fig:shape_functions}
\end{figure}
Despite having the same ${c_{1}}$ and ${c_{2}}$ for both Example~\ref{ex:2} and Example~\ref{ex:3}, their shape functions behave differently as shown in Figure~\ref{fig:shape_functions}.
Moreover, the user can select the amount of nonconvexity added to the surrogate loss function ${\rho_{\mu}(\cdot)}$ by choosing the constant ${c}$ in \eqref{eq:continuation}.
The shape function in Example~\ref{ex:2}, represented by the orange line, approaches ${f(\cdot)=\alpha^{\star}}$ faster than the one in Example~\ref{ex:3}, represented by the green line, which may lead to faster convergence at the expense of robustness of the solver.
Based on the problems tested in this paper, Example~\ref{ex:3} is found to be the most robust among the three examples.
However, as stated earlier, a detailed analysis as to what shape function performs the best is not provided.
As such, the choice of a shape function is left to the user.


\subsection{Adaptive GNC Weight Update}
\label{sec:methodology_weightupdate}

Similar to Yang \etal's work \cite{Yang2020GNC}, \eqref{eq:brd_gnc} is optimized using IRLS with changing ${\mu}$.
The states ${\mbf{x}}$ and the weights ${w_{i}}$ are optimized alternatively and iteratively.
Recall that the Mahalanobis distance of the residual is a function of ${\mbf{x}}$.
Thus, minimizing \eqref{eq:brd_gnc} with respect to ${\mbf{x}}$ and fixed weights ${w_{i}}$ results in minimizing \eqref{eq:IRLS} because the outlier process function does not depend on ${\mbf{x}}$.
On the other hand, minimizing \eqref{eq:brd_gnc} with respect to ${w_{i}}$ and fixed ${\mbfbar{x}}$ results in
\begin{align}
    \mbf{w}^{\star} = \underset{w_{i}\in(0,1]}{\argmin} \sum_{i=1}^{N} \onehalf w_{i} \epsilon_{i}\left(\mbfbar{x}\right)^{2} + \Phi_{\rho_{\mu}}\left(w_{i}\right).
    \label{eq:agnc_weightoptimization}
\end{align}
The solution to \eqref{eq:agnc_weightoptimization} for changing values of ${\mu}$ is given as
\begin{align}
    w^{\star}\left(\epsilon, \alpha^{\star}\right) & =
    \begin{cases}
        1                                                                      & \textrm{if } f = 2,       \\
        \f{2}{\epsilon^{2} + 2}                                                & \textrm{if } f = 0,       \\
        \exp\left(-\onehalf\epsilon^{2}\right)                                 & \textrm{if } f = -\infty, \\
        \left( \f{\epsilon^{2}}{\abs{f(\mu, \alpha^{\star}) - 2}} + 1 \right)^{\nicefrac{f(\mu, \alpha^{\star})}{2} - 1} & \textrm{otherwise}.
    \end{cases}
    \label{eq:agnc_weights}
\end{align}

\subsection{Adaptive GNC Algorithm}
\label{sec:methodology_agnc}
\begin{algorithm}[tb]
    \caption{Implementation of adaptive GNC}\label{alg:gnc-alf}
    \begin{algorithmic}
        \Require ${\mbs{\epsilon} = \bbm \epsilon_{1} & \epsilon_{2} & \ldots & \epsilon_{N} \ebm^{\trans}}$, ${f(\mu, \alpha^{\star})}$
        \State Compute ${\alpha^{\star}}$ using \eqref{eq:cheb_alphastar} or select manually.
        \State Select an initial convexity parameter ${\mu \mid f(\mu, \alpha^{\star}) \to 2}$
        \While{True}
        \State Compute ${f(\mu, \alpha^{\star})}$.
        \State Compute the optimal weights ${\mbf{w}^{\star}}$ for ${\mbs{\epsilon}}$ using \eqref{eq:agnc_weights}.
        \State Solve for the states ${\mbf{x}^{\star}}$ with ${\mbf{w}^{\star}}$ using \eqref{eq:IRLS}.
        \If{Not converged}
        \State Update ${\mu}$ such that ${f(\mu, \alpha^{\star})}$ approaches $\alpha^{\star}$.
        \EndIf
        \If{${f \approx \alpha^{\star}}$}
        \State Compute a new ${\alpha^{\star}}$ using \eqref{eq:cheb_alphastar}.
        \State Set ${\mu}$ to the initial value.
        \EndIf
        \EndWhile
    \end{algorithmic}
\end{algorithm}

This section describes the implementation of the proposed algorithm shown in Algorithm~\ref{alg:gnc-alf}.
First, a shape function ${f(\cdot)}$ is selected by the user.
The choice of ${f(\cdot)}$ does not affect the ultimate shape of the optimal loss function but only affects the rate of nonconvexity added at each iteration.

Given residuals ${\mbs{\epsilon} = \bbm \epsilon_{1} & \ldots & \epsilon_{N} \ebm^{\trans}}$, the optimal shape parameter ${\alpha^{\star}}$ is computed using \eqref{eq:cheb_alphastar}.
Also, the convexity parameter ${\mu}$ is set such that ${f(\mu, \alpha^{\star}) \lesssim 2}$ to make the surrogate loss function ${\rho_{\mu}(\cdot)}$ convex.
With ${\alpha^{\star}}$ and ${\mu}$, the optimal weights ${\mbf{w}^{\star}}$ are computed using \eqref{eq:agnc_weights}, followed by computing the states ${\mbf{x}^{\star}}$ using \eqref{eq:IRLS} with the optimal weights ${\mbf{w}^{\star}}$.
If the optimization has not converged, ${\mu}$ is updated to add nonconvexity to the surrogate cost function.
This process is repeated either until the optimization converges or the nonconvexity is saturated, where saturation is defined as ${f \approx \alpha^{\star}}$.
If the nonconvexity is saturated, then a new optimal shape parameter ${\alpha^{\star}}$ is computed using \eqref{eq:cheb_alphastar} and the process is repeated from the initial ${\mu}$.

Although the algorithm computes the optimal shape parameter ${\alpha^{\star}}$ based on the residual distribution ${q(\mbs{\epsilon})}$, the optimal shape parameter ${\alpha^{\star}}$ can be selected by the user.
For example, if the user wants to use the Cauchy or Geman-McClure loss function, the user can select ${\alpha^{\star}=0}$ or ${\alpha^{\star}=-2}$, respectively.
Ideally, the adaptive shape parameter should perform better or on par with the user-selected shape parameter.
The resulting function, GNC-ADAPT, is the first contribution of this paper.

\subsection{GNC Algorithm with AMB Loss}
\label{sec:methodology_gncamb}

Recall that most of the robust loss functions that are subsets of \eqref{eq:barron_adaptive_loss} assume a Gaussian-like residual distribution, and this assumption can inadvertently lead to problems with the weighting scheme \cite{Hitchcox2022Mind}.
Thus, the adaptive GNC loss function is extended to incorporate the ``mode gap'' approach of Hitchcox and Forbes \cite{Hitchcox2022Mind}.

The mode of the residual distribution ${\tilde{\epsilon}}$ is estimated using \eqref{eq:mb_mode} by fitting an MB distribution \eqref{eq:pmb} to the residual distribution ${q(\epsilon)}$ as shown in \eqref{eq:a_star}.
With the mode of the residual distribution, the mode-shifted residuals and truncation bound are computed as
\begin{align}
    \label{eq:mode_shifted_residuals}
    \mbs{\xi}    & = \mbs{\epsilon} - \tilde{\epsilon}\mathbbm{1}, \\
    \label{eq:mode_shifted_truncation}
    \tilde{\tau} & = \tau - \tilde{\epsilon},
\end{align}
where ${\mathbbm{1}}$ is a column matrix of ones.
With the mode-shifted residuals ${\mbs{\xi}}$, the optimal shape parameter ${\alpha^{\star}}$ is computed using \eqref{eq:amb_alpha} and \eqref{eq:amb_partition}.
Then, Algorithm~\ref{alg:gnc-alf} is performed on the mode-shifted residuals ${\mbs{\xi}}$ with the optimal shape parameter ${\alpha^{\star}}$.
The resulting loss function, GNC-AMB, is the second contribution of this paper.

\section{Results}
\label{sec:results}

The main contribution of this paper is the application of GNC to an adaptive loss function.
This is expected to be as or more robust to outliers than GNC with a particular loss function, and as or more robust to initialization errors than non-GNC adaptive loss functions.
The developed method is tested against GNC using loss functions Cauchy, GM, and TLS, as well as the adaptive loss functions from \cite{Barron2019General, Chebrolu2021Adaptive}.
These loss functions are chosen for their good performance in a similar study \cite{Babin2019Analysis, Hitchcox2022Mind}.
Note that Cauchy and GM loss functions are fused with GNC by selecting a fixed value of ${\alpha}$, ${0}$ and ${-2}$ respectively, in the generation equation, and TLS is proposed by Yang \etal \cite{Yang2020GNC}.

The proposed method is tested in three experiments: point-cloud alignment, mesh registration, and pose graph optimization.
All experiments assess the performance of the proposed method in comparison to the state-of-the-art GNC loss functions as well as the non-GNC counterparts.

\begin{table}[tb]
    \renewcommand{\arraystretch}{1.25} 
    \centering
    \caption{
        Abbreviations used for different loss functions.
    }
    \label{tab:abbrev}
    \begin{tabular}{ll}
        \toprule
        Method        & Description                 \\
        \hline
        Cauchy        & \cite{Black1996Robust}      \\
        Geman-McClure & \cite{Geman1986Baysian}     \\
        Adaptive loss & \cite{Chebrolu2021Adaptive} \\
        Adaptive MB   & \cite{Hitchcox2022Mind}     \\
        GNC-Cauchy    & GNC with Cauhcy             \\
        GNC-GM        & GNC with GM                 \\
        GNC-ADAPT     & GNC with adaptive loss      \\
        GNC-AMB       & GNC with Adaptive MB        \\
        \bottomrule
    \end{tabular}
\end{table}

\subsection{Point-Cloud Alignment}
\label{sec:results_pca}

\begin{figure*}[tb]
    \centering
    \begin{subfigure}[t]{0.32\linewidth}
        \centering
        \includegraphics[width=\linewidth]{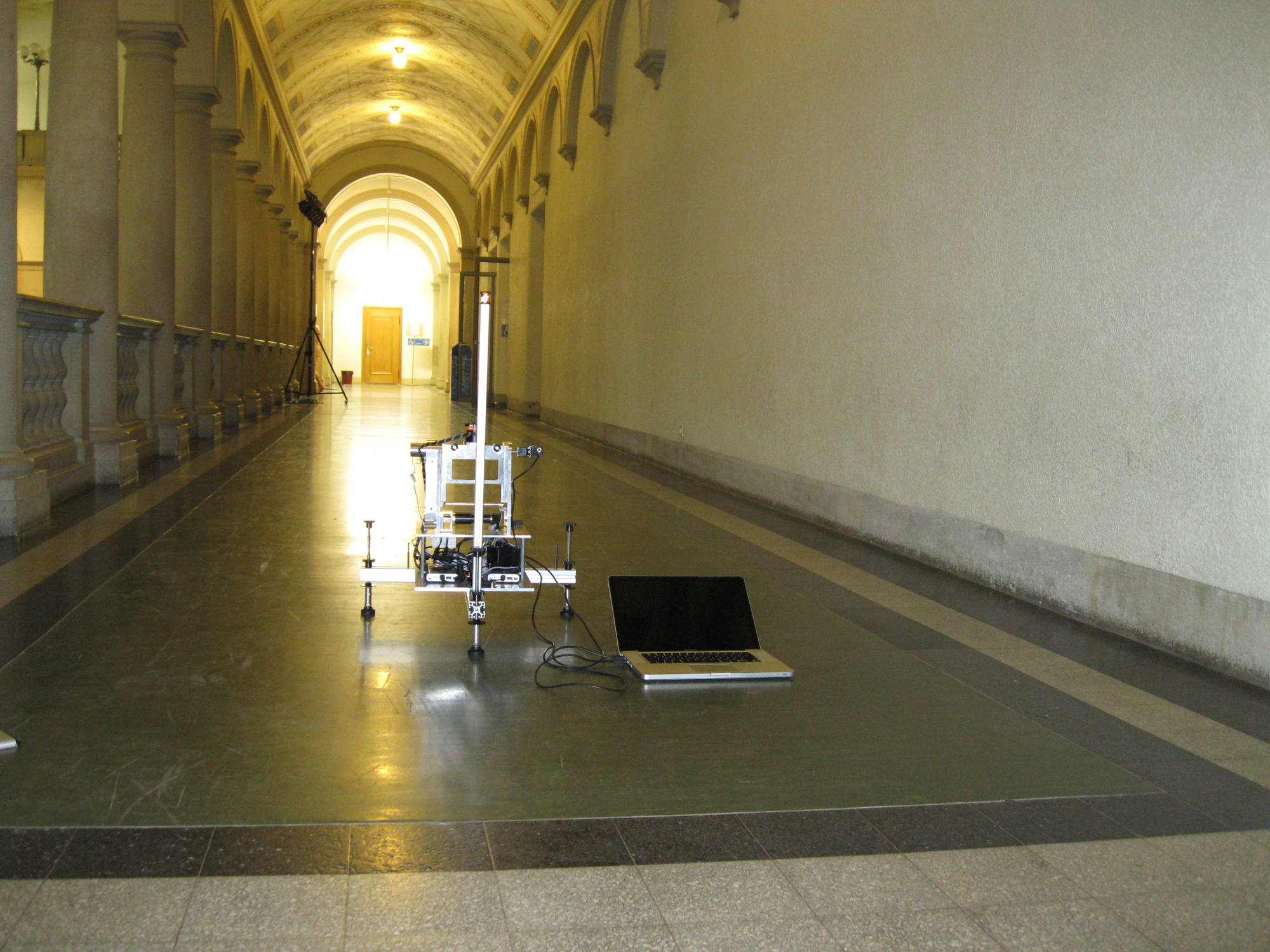}
        \caption{ETH Hauptgebaude (EH)}
        \label{fig:EH}
    \end{subfigure}
    \begin{subfigure}[t]{0.32\linewidth}
        \centering
        \includegraphics[width=\linewidth]{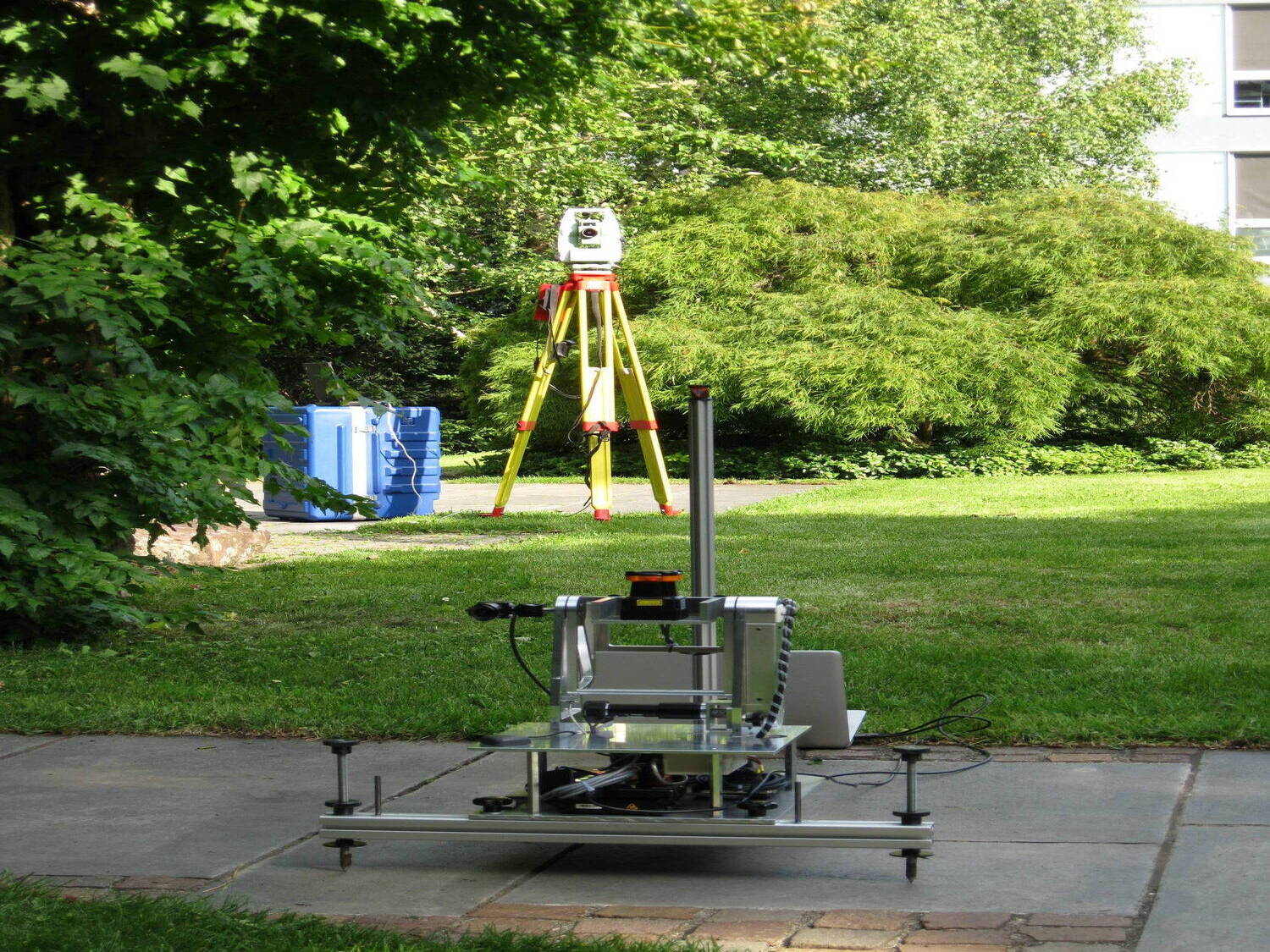}
        \caption{Gazebo in Summer (GZ)}
        \label{fig:GZ}
    \end{subfigure}
    \begin{subfigure}[t]{0.32\linewidth}
        \centering
        \includegraphics[width=\linewidth]{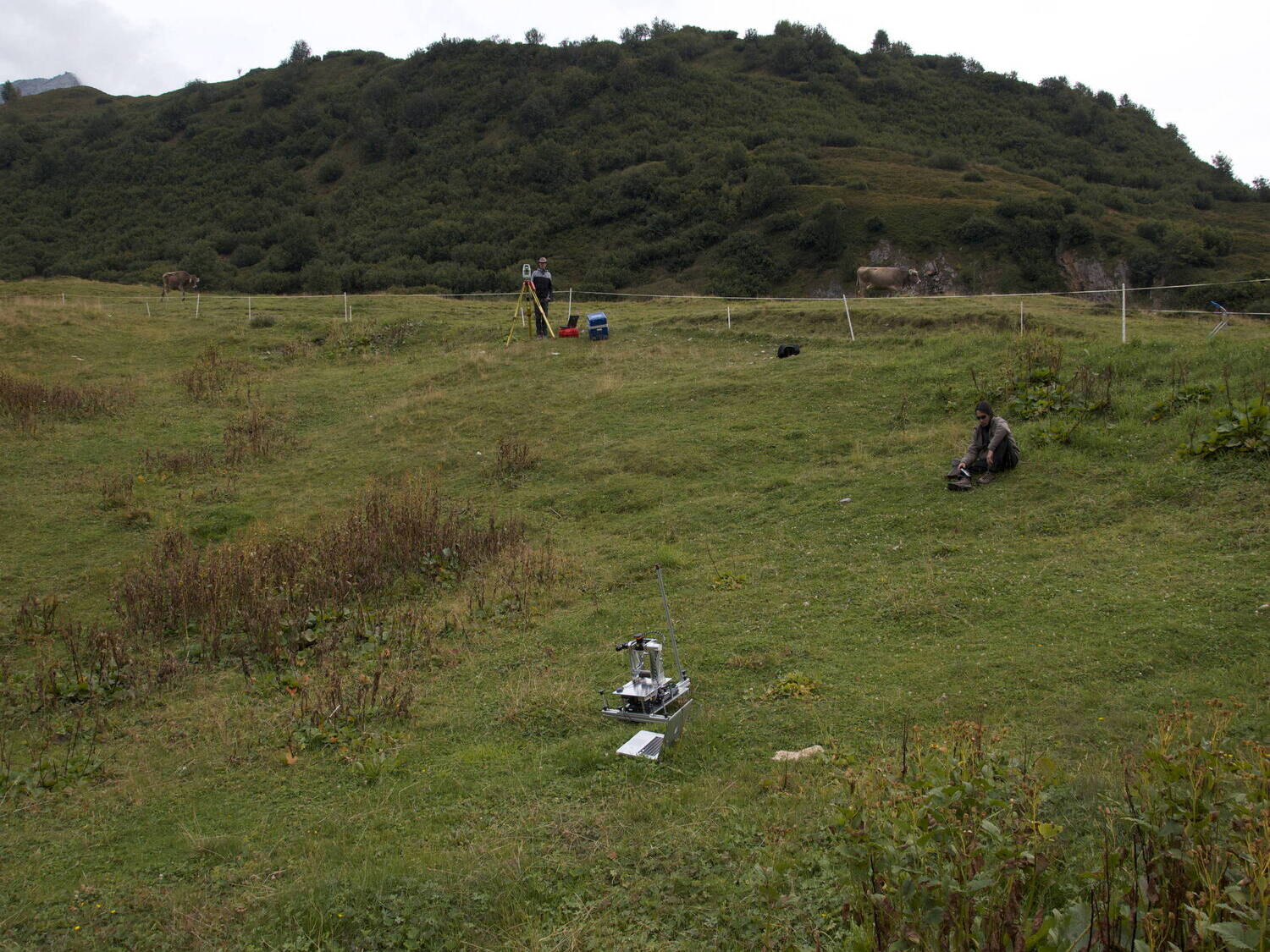}
        \caption{Mountain plain (MP)}
        \label{fig:MP}
    \end{subfigure}
    \caption{Contextual photographs of different environments~\cite{Pomerleau2012Challenging}.}
    \label{fig:pca_dataset}
\end{figure*}
\begin{figure*}[tb]
    \centering
    \begin{subfigure}[t]{0.32\linewidth}
        \centering
        \includegraphics[width=\linewidth]{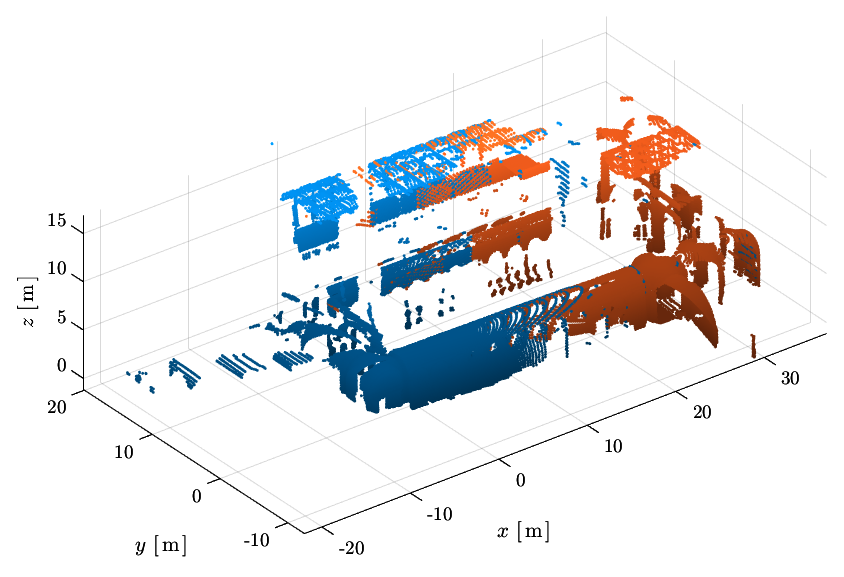}
        \caption{30\% scan overlap}
        \label{fig:overlap_30}
    \end{subfigure}
    \begin{subfigure}[t]{0.32\linewidth}
        \centering
        \includegraphics[width=\linewidth]{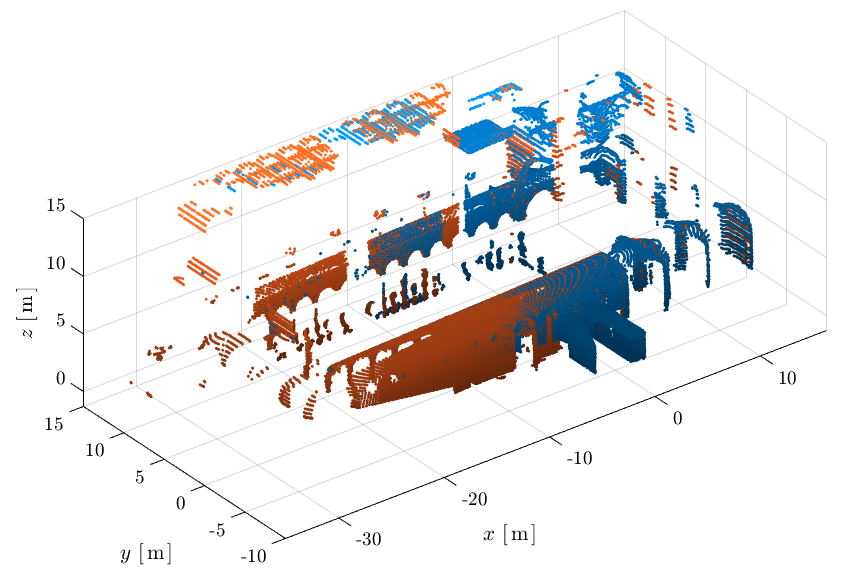}
        \caption{50\% scan overlap}
        \label{fig:overlap_50}
    \end{subfigure}
    \begin{subfigure}[t]{0.32\linewidth}
        \centering
        \includegraphics[width=\linewidth]{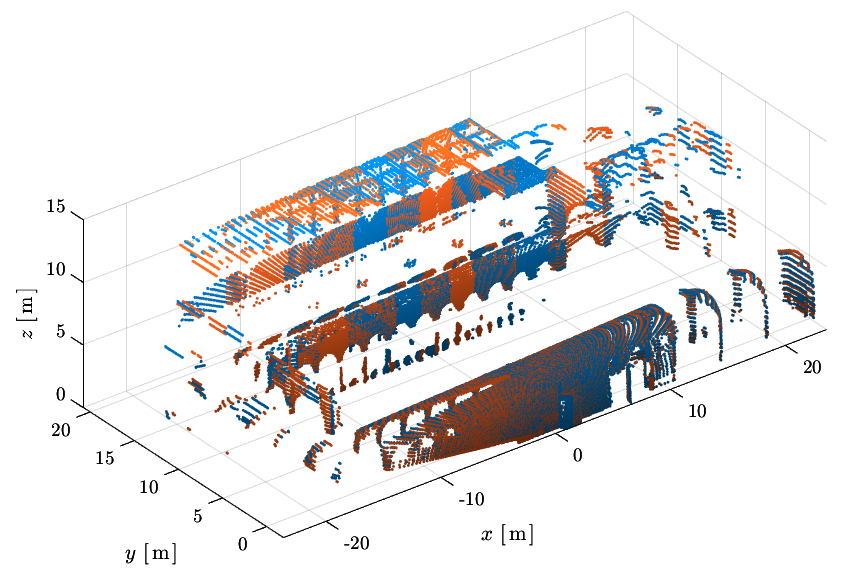}
        \caption{90\% scan overlap}
        \label{fig:overlap_90}
    \end{subfigure}
    \caption{Three point-cloud scan pairs at different overlap ratios from ETH Hauptgebaude dataset.}
    \label{fig:overlap}
\end{figure*}

Point-cloud alignment is an algorithm used to find a 3D rigid transformation that best aligns two input point clouds.
This is a fundamental problem in robotics and is used in applications such as simultaneous localization and mapping (SLAM), object recognition, and 3D reconstruction.
The two point clouds are a fixed point cloud, ${\mc{P}=\{\mbf{r}^{p_{i}s_{1}}_{\ell_{1}}\}^{N}_{i=1}}$, and a moving point cloud, ${\mc{Q}=\{\mbf{r}^{q_{i}s_{2}}_{\ell_{2}}\}^{M}_{i=1}}$.
The point measurement ${\mbf{r}^{p_{i}s_{1}}_{\ell_{1}}}$ describes the position of the ${i}$-th point ${p_{i}}$ relative to the position of the sensor ${s_{1}}$ resolved in the sensor frame ${\mc{F}_{\ell_{1}}}$ at time ${t_{1}}$.
The relative transformation is represented as an element of matrix Lie group ${SE(3)}$,
\begin{align}
    \mbf{T}^{s_{2}s_{1}}_{\ell_{1}\ell_{2}} & = \bbm \mbf{C}_{\ell_{1}\ell_{2}} & \mbf{r}^{s_{2}s_{1}}_{\ell_{1}} \\ \mbf{0} & 1 \ebm \in SE(3),
\end{align}
where ${\mbf{C} \in SO(3) = \{\mbf{C} \in \mathbb{R}^{3\times3} \mid \mbf{C}\mbf{C}^{\trans} = \mbf{1}, \det \mbf{C} = +1\}}$ is the direction cosine matrix that represents the attitude of the laser frame at ${t_{1}}$ with respect to the laser frame at ${t_{2}}$, and ${\mbf{r}^{s_{2}s_{1}}_{\ell_{1}} \in \mathbb{R}^{3}}$ represents the position of the sensor at time ${t_{2}}$ relative to the sensor at time ${t_{1}}$, resolved in the sensor frame at time ${t_{1}}$.
For readability, ${\mbf{T}_{12} = \mbf{T}^{s_{2}s_{1}}_{\ell_{1}\ell_{2}}}$.
\begin{table}[tb]
    \renewcommand{\arraystretch}{1.25} 
    \centering
    \caption{ICP experiment setting.}
    \label{tab:icp_setting}
    \begin{tabular}{ll}
        \toprule
        Downsample                   & \texttt{VoxelGrid} with $\SI{10}{\centi\meter}$ voxel size \\
        Normals                      & 15 nearest neighbours                                      \\
        Point association            & Single nearest neighbour                                   \\
        Residual                     & Point-to-plane error                                       \\
        \multirow{2}{*}{Convergence} & ${\norm{\delta\mbs{\phi}^{\star}} < \SI{1e-3}{\radian}}$   \\
                                     & ${\norm{\delta\mbs{\rho}^{\star}} < \SI{1e-3}{\meter}}$    \\
        Maximum iterations           & 50                                                         \\
        \bottomrule
    \end{tabular}
\end{table}

The iterative closest point (ICP) algorithm is used to solve a point-cloud alignment problem
subject to outliers and initialization errors.
The ICP algorithm, which solves a nonconvex optimization problem from a prior relative transformation ${\mbfcheck{T}_{12}}$, first associates each point in ${\mc{P}}$ to its nearest Euclidean neighbor in ${\mc{Q}}$ using ${\mbfcheck{T}_{12}}$.
The associated points in ${\mc{Q}}$ are now denoted as ${\tilde{\mc{Q}}=\{\mbf{r}^{q_{i}s_{2}}_{\ell_{2}}\}^{N}_{i=1}}$ where ${N}$ is the total number of point correspondences.
Then, the optimization problem
\begin{align}
    \mbf{T}_{12}^{\star} = \underset{\mbf{T}_{12} \in SE(3)}{\argmin} \sum_{i=1}^{N} \f{1}{2} w_{i} \norm{\mbf{e}_{i}\left(\mbf{T}_{12}, \mbf{r}^{p_{i}s_{1}}_{\ell_{1}}, \mbf{r}^{q_{i}s_{2}}_{\ell_{2}}\right)}_{\mbs{\Sigma}^{-1}_{i}}^{2}
    \label{eq:icp}
\end{align}
is solved, where ${w_{i}\in(0, 1]}$ are association weights, ${\mbf{e}_{i}}$ are association errors, and ${\mbs{\Sigma}_{i}}$ are association error covariances.
Then, the point correspondences are found again with the updated relative transformation.
This process is repeated until a convergence criterion is met.
The ICP problem is solved both using traditional loss functions and their GNC variants, including the proposed GNC-ADAPT and GNC-AMB methods from Section~\ref{sec:methodology}.
The point-cloud alignment experiment evaluates the convergence and success rates, alignment errors, and processing times in different initial perturbation levels, overlap ratios, and structuredness of the environment.

\subsubsection{Experiment Setting}
An open-source point-cloud dataset published by Pomerleau \etal is used in this experiment \cite{Pomerleau2012Challenging}.
The dataset contains eight point-cloud sequences where each sequence contains ground-truth poses, point-cloud scans, and the overlap ratio between each scan pair. 
Recall that naively solving \eqref{eq:icp} can lead to a poor estimate in the presence of outlier correspondences.
Outlier correspondences are frequently found in the case of poor ICP initialization, or when there is a low overlap ratio between ${\mc{P}}$ and ${\mc{Q}}$.
Thus, to thoroughly evaluate the proposed loss functions, ICP performance is evaluated subject to three input settings: the environment diversity, the overlap ratio, and the initial perturbation.

Three point-cloud datasets are chosen from \cite{Pomerleau2012Challenging}, representing a wide range in the diversity of the scanned environment.
``ETH Hauptgebaude'' (EH) is chosen to represent structured scans, while ``Gazebo in Summer'' (GZ) and ``Mountain Plain'' (MP) are chosen to represent semi-structured and unstructured scans, respectively.
The ``structured'' level is a qualitative assessment of the number of geometric primitives visible in the scene.
The datasets are shown in Figure~\ref{fig:pca_dataset}.

For each point-cloud sequence, two scans are randomly selected to represent ${\mc{P}}$ and ${\mc{Q}}$, and the selected scan pair is assigned to the corresponding overlap ratio bins.
The bins span from 30\% to 90\% with a bin size of 20\%.
An example of scan pairs with different overlap ratios is shown in Figure~\ref{fig:overlap}.

For each scan pair, three levels of initial relative transformation difficulty are tested: ``low,'' ``medium,'' and ``high.''
The initial relative transformation is computed using ${\mbfcheck{T}_{12} = \mbf{T}_{12}\delta\mbfcheck{T}}$, where ${\mbf{T}_{12}}$ is the ground-truth transformation and ${\delta\mbfcheck{T}}$ is the initial perturbation defined in the matrix Lie algebra ${\mathfrak{se}(3)}$ \cite{Arsenault2020Practical},
\begin{align}
    \log\left(\delta\mbfcheck{T}\right) = \delta\mbscheck{\xi}^{\wedge} = \bbm \delta\mbscheck{\phi} \\ \delta\mbscheck{\rho} \ebm^{\wedge} \in \mathfrak{se}(3),
    \label{eq:perturbation}
\end{align}
where ${(\cdot)^{\wedge}:\mathbb{R}^{6}\to\mathfrak{se}(3)}$ is an isometric operator, and ${\delta\mbs{\phi}}$ and ${\delta\mbs{\rho}}$ represent the perturbation in rotation and translation in ${\mathfrak{se}(3)}$, respectively.
Thus, the perturbation $\delta\mbfcheck{T}$ is composed of
\begin{align}
    \delta\mbscheck{\phi} \sim \mc{N}\left(\mbf{0}, \sigma_{\phi}^{2}\mbf{1}\right), \quad \delta\mbscheck{\rho} \sim \mc{N}\left(\mbf{0}, \sigma_{\rho}^{2}\mbf{1}\right),
    \label{eq:initial_perturbation}
\end{align}
where ${\sigma_{\phi}}$ and ${\sigma_{\rho}}$ are the perturbation standard deviations in the attitude and position, respectively.
The initial perturbation difficulties are based on Table~4 of \cite{Pomerleau2013Comparing}.
The standard deviations in \cite{Pomerleau2013Comparing} are chosen such that ${\|\delta\mbscheck{\phi}\| < \delta\phi_{\max}}$ and ${\|\delta\mbscheck{\rho}\| < \delta\rho_{\max}}$, where ${\delta\phi_{\max}}$ is ${\SI{10}{\degree}}$, ${\SI{20}{\degree}}$, and ${\SI{45}{\degree}}$, and ${\delta\rho_{\max}}$ is ${\SI{0.1}{\meter}}$, ${\SI{0.5}{\meter}}$, and ${\SI{1}{\meter}}$, for easy, medium, and hard levels, respectively.

For each point-cloud sequence, 20 scan pairs are tested for each perturbation level and scan overlap ratio setting.
With 3 point-cloud sequences, 3 scan overlap ratios, and 3 initial perturbation levels, a total of 540 trials are conducted for each of the 10 robust kernels from Table~\ref{tab:abbrev}.
ICP experiment setting is given in Table~\ref{tab:icp_setting}.
For the proposed methods, the shape function from Example~\ref{ex:3} is used.

\begin{figure*}[tb]
    \centering
    \includegraphics[width=\linewidth]{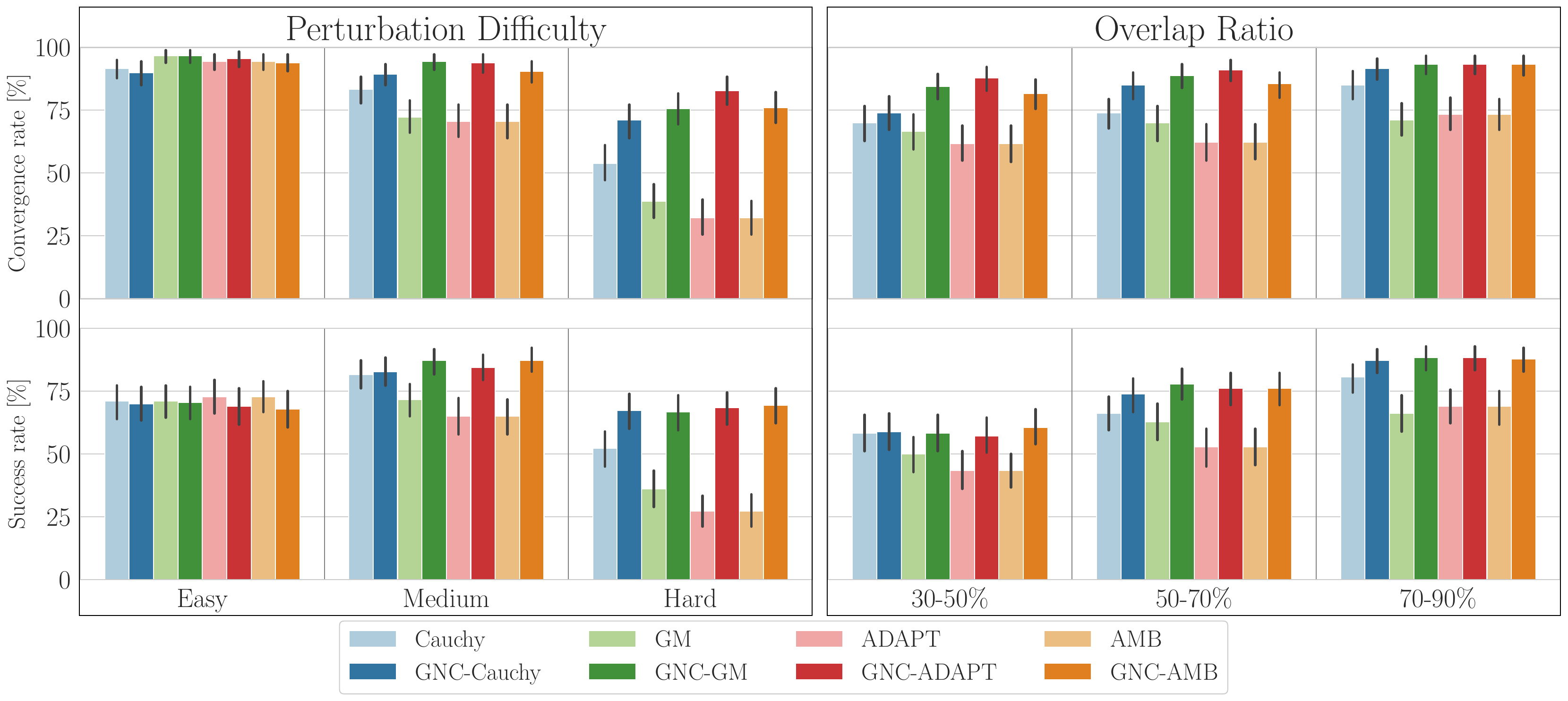}
    \caption{
        The percentage of ICP convergence and success in different initial perturbation levels (left) and overlap ratios (right) are shown here for each loss function (transparent) and its GNC variant (opaque).
        ICP is considered successful if both the final rotation and translation errors are less than the initial perturbation, ${\|\delta\mbshat{\phi}\| < \|\delta\mbscheck{\phi}\|}$ and ${\|\delta\mbshat{\rho}\| < \|\delta\mbscheck{\rho}\|}$, and the total number of iterations is less than the maximum specified in Table~\ref{tab:icp_setting}.
        The GNC variants show higher convergence and success rates than the traditional loss functions in all initial perturbation levels and overlap ratios.
    }
    \label{fig:pca_percent_success}
\end{figure*}
The association error is defined as the point-to-plane error \cite{Chen1991Object}.
However, the distribution of point-to-point error \cite{Besl1992Method} is used to compute the shape parameters for the adaptive loss functions such that the distance between two associated points is considered when rejecting outlier matches.
The point-to-point error and its squared Mahalanobis are defined as
\begin{align}
    \mbf{e}_{i}      & = \mbf{r}^{p_{i}s_{1}}_{\ell_{1}} - \left(\mbf{C}_{\ell_{1}\ell_{2}}\mbf{r}^{q_{i}s_{2}}_{\ell_{2}} + \mbf{r}^{s_{2}s_{1}}_{\ell_{1}}\right), \\
    \epsilon_{i}^{2} & = \mbf{e}_{i}^{\trans}\mbs{\Sigma}_{i}^{-1}\mbf{e}_{i},
\end{align}
where ${\mbs{\Sigma}_{i} = \mbf{C}_{\ell_{1}\ell_{2}}\mbf{R}_{p_{i}}\mbf{C}_{\ell_{1}\ell_{2}}^{\trans} + \mbf{R}_{q_{i}}}$ is the covariance matrix of the ${i}$-th point-to-point error, computed using ${\mbf{R}_{p_{i}}}$ and ${\mbf{R}_{q_{i}}}$, the covariance on the measured points ${p_{i}}$ and ${q_{i}}$, respectively.
The covariance of ${p_{i}}$ is set to ${\mbf{R}_{p_{i}} = \mbf{R}_{q_{i}} = \sigma^{2}_{\ell}\mbf{1}}$, where ${\sigma_{\ell}=\SI{3}{\centi\metre}}$ as reported in \cite{Pomerleau2012Challenging}.

\subsubsection{Convergence and Success Rate}
ICP is considered to have converged when the incremental update ${\delta\mbs{\xi}^{\star}}$ is less than the convergence thresholds defined in Table~\ref{tab:icp_setting} before the maximum number of iterations is reached.
Additionally, each trial is considered ``successful'' when both posterior attitude error ${\delta\mbshat{\phi}}$ and the posterior position error ${\delta\mbshat{\rho}}$ are less than the initial perturbations, and the total number of iterations is less than the maximum threshold.

The top row of Figure~\ref{fig:pca_percent_success} shows the convergence rate of each method at different difficulty levels (left) and at different overlap ratios (right), whereas the bottom row shows the corresponding success rate.
The GNC variants, derived using a single equation \eqref{eq:rlf_surrogate}, show higher convergence and success rates than the traditional loss functions in all initial perturbation levels and overlap ratio, showing robustness to both outliers and initial conditions.
Notice the difference in performance between the baseline functions and their GNC counterparts becomes more pronounced as the initial perturbation increases but not as much for increasing overlap ratio.
This shows that incorporating GNC is more significant in the presence of large initial perturbations.

Note that the success rate for the ``easy'' initial perturbations is worse than medium difficulty.
Recall that a ``successful'' trial is determined by the posterior error being less than the initial perturbation.
For ``easy'' initial perturbations, the distribution on the posterior errors in Figure~\ref{fig:pca_percent_success} (light blue) is not significantly different from the distribution on the prior perturbation (dark blue), compared to the difference seen for ``medium'' initial perturbations (orange).
\begin{figure}[tb]
    \centering
    \includegraphics[width=\linewidth]{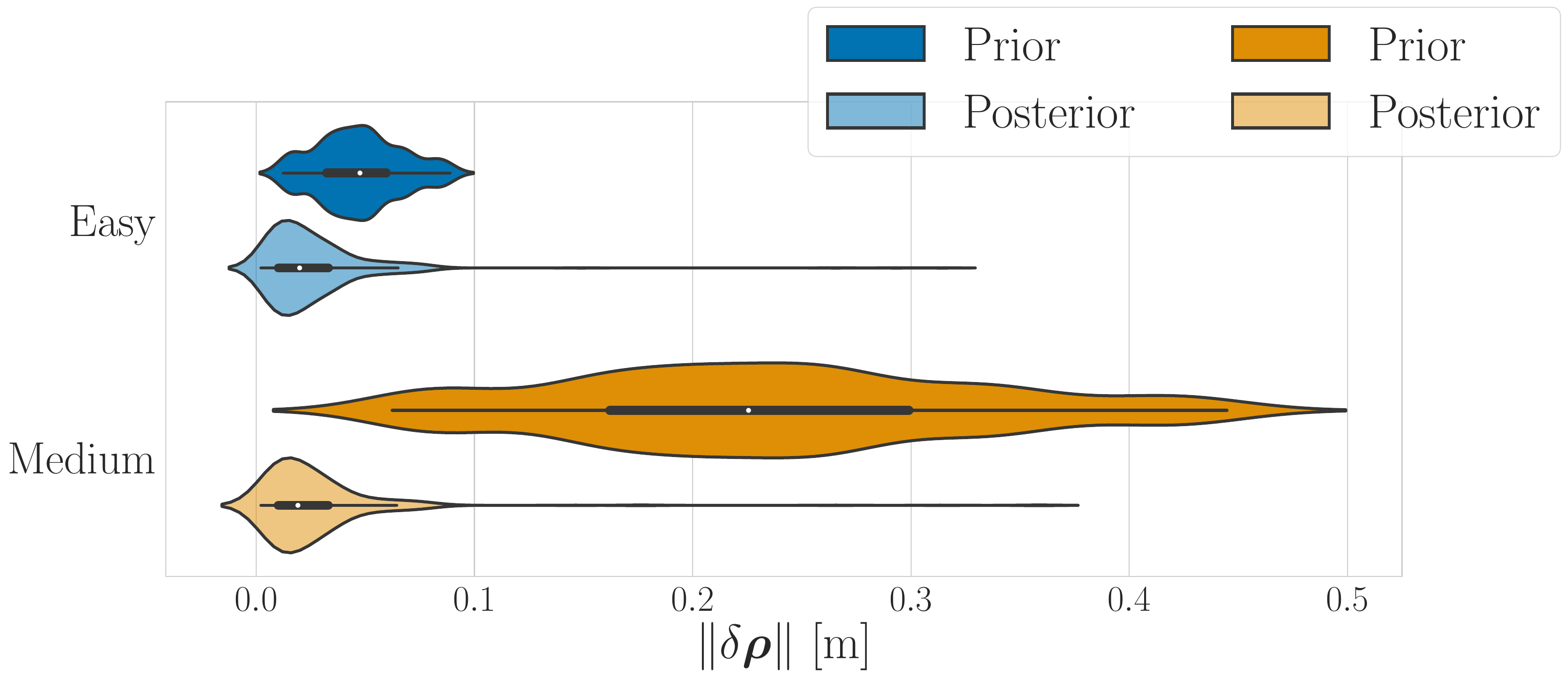}
    \caption{Prior and posterior translation error distribution after convergence for easy and medium difficulty levels for all trials with EH dataset.}
    \label{fig:prior_posterior}
\end{figure}

\subsubsection{Alignment Errors}
\begin{table*}[tb]
    \renewcommand{\arraystretch}{1.25} 
    \centering
    \caption{
        The proposed methods are compared against GNC-TLS and GNC-GM proposed by Yang \etal in \cite{Yang2020GNC} in terms of the alignment errors and processing time.
        The results are broken down by environment and difficulty level.
        The errors are reported as 50\%-\textcolor{Maroon}{75\%}-\textcolor{NavyBlue}{90\%} percentiles.
        The lowest error for each percentile for every dataset and difficulty in each row is highlighted in bold.
    }
    \label{tab:icp_results}
    \begin{tabular}{l|cc|ccccc}
        \toprule
        Metric                                                         & Dataset             & Difficulty & GNC-TLS \cite{Yang2020GNC}                                          & GNC-GM \cite{Yang2020GNC}                                                             & GNC-GM (from proposed)                                                       & GNC-ADAPT                                                                             & GNC-AMB                                                                               \\ \hline
        \multirow{9}{*}{$\norm{\delta\mbshat{\phi}}$ [$\mathrm{deg}$]} & \multirow{3}{*}{EH} & Easy       & 0.175-\textcolor{Maroon}{\textbf{0.236}}-\textcolor{NavyBlue}{0.302} & 0.167-\textcolor{Maroon}{0.241}-\textcolor{NavyBlue}{\textbf{0.301}}                   & 0.165-\textcolor{Maroon}{0.244}-\textcolor{NavyBlue}{0.302}                   & \textbf{0.160}-\textcolor{Maroon}{0.246}-\textcolor{NavyBlue}{0.302}                   & 0.169-\textcolor{Maroon}{0.244}-\textcolor{NavyBlue}{0.303}                            \\
                                                                       &                     & Medium     & 0.178-\textcolor{Maroon}{0.261}-\textcolor{NavyBlue}{0.308}          & 0.179-\textcolor{Maroon}{0.256}-\textcolor{NavyBlue}{0.311}                            & 0.175-\textcolor{Maroon}{0.256}-\textcolor{NavyBlue}{0.311}                   & \textbf{0.171}-\textcolor{Maroon}{\textbf{0.248}}-\textcolor{NavyBlue}{\textbf{0.305}} & 0.183-\textcolor{Maroon}{0.261}-\textcolor{NavyBlue}{0.305}                            \\
                                                                       &                     & Hard       & \textbf{0.143}-\textcolor{Maroon}{0.217}-\textcolor{NavyBlue}{0.337} & 0.143-\textcolor{Maroon}{0.207}-\textcolor{NavyBlue}{0.326}                            & 0.144-\textcolor{Maroon}{\textbf{0.207}}-\textcolor{NavyBlue}{0.330}          & 0.145-\textcolor{Maroon}{0.208}-\textcolor{NavyBlue}{0.328}                            & 0.143-\textcolor{Maroon}{0.217}-\textcolor{NavyBlue}{\textbf{0.323}}                   \\
        \cline{2-8}                                                    & \multirow{3}{*}{GZ} & Easy       & 0.335-\textcolor{Maroon}{0.454}-\textcolor{NavyBlue}{0.575}          & 0.322-\textcolor{Maroon}{0.477}-\textcolor{NavyBlue}{0.558}                            & 0.325-\textcolor{Maroon}{0.475}-\textcolor{NavyBlue}{0.557}                   & 0.323-\textcolor{Maroon}{0.451}-\textcolor{NavyBlue}{0.540}                            & \textbf{0.305}-\textcolor{Maroon}{\textbf{0.392}}-\textcolor{NavyBlue}{\textbf{0.523}} \\
                                                                       &                     & Medium     & 0.370-\textcolor{Maroon}{0.555}-\textcolor{NavyBlue}{0.715}          & \textbf{0.342}-\textcolor{Maroon}{0.519}-\textcolor{NavyBlue}{0.673}                   & 0.357-\textcolor{Maroon}{0.520}-\textcolor{NavyBlue}{\textbf{0.666}}          & 0.358-\textcolor{Maroon}{\textbf{0.515}}-\textcolor{NavyBlue}{0.675}                   & 0.345-\textcolor{Maroon}{0.538}-\textcolor{NavyBlue}{0.687}                            \\
                                                                       &                     & Hard       & 0.405-\textcolor{Maroon}{0.550}-\textcolor{NavyBlue}{9.292}          & 0.367-\textcolor{Maroon}{0.499}-\textcolor{NavyBlue}{0.571}                            & 0.342-\textcolor{Maroon}{0.491}-\textcolor{NavyBlue}{0.555}                   & 0.343-\textcolor{Maroon}{\textbf{0.455}}-\textcolor{NavyBlue}{2.023}                   & \textbf{0.331}-\textcolor{Maroon}{0.473}-\textcolor{NavyBlue}{\textbf{0.522}}          \\
        \cline{2-8}                                                    & \multirow{3}{*}{MP} & Easy       & 0.247-\textcolor{Maroon}{0.352}-\textcolor{NavyBlue}{\textbf{0.419}} & 0.228-\textcolor{Maroon}{0.328}-\textcolor{NavyBlue}{0.433}                            & 0.226-\textcolor{Maroon}{\textbf{0.288}}-\textcolor{NavyBlue}{0.426}          & \textbf{0.219}-\textcolor{Maroon}{0.331}-\textcolor{NavyBlue}{0.452}                   & 0.238-\textcolor{Maroon}{0.378}-\textcolor{NavyBlue}{0.449}                            \\
                                                                       &                     & Medium     & 0.320-\textcolor{Maroon}{0.412}-\textcolor{NavyBlue}{0.530}          & 0.340-\textcolor{Maroon}{0.436}-\textcolor{NavyBlue}{0.537}                            & \textbf{0.298}-\textcolor{Maroon}{0.411}-\textcolor{NavyBlue}{\textbf{0.475}} & 0.307-\textcolor{Maroon}{\textbf{0.391}}-\textcolor{NavyBlue}{0.502}                   & 0.338-\textcolor{Maroon}{0.446}-\textcolor{NavyBlue}{0.529}                            \\
                                                                       &                     & Hard       & 0.309-\textcolor{Maroon}{0.480}-\textcolor{NavyBlue}{11.41}          & 0.282-\textcolor{Maroon}{0.468}-\textcolor{NavyBlue}{\textbf{0.740}}                   & \textbf{0.274}-\textcolor{Maroon}{\textbf{0.445}}-\textcolor{NavyBlue}{0.827} & 0.316-\textcolor{Maroon}{0.538}-\textcolor{NavyBlue}{8.597}                            & 0.288-\textcolor{Maroon}{0.488}-\textcolor{NavyBlue}{0.896}                            \\ \hline
        \multirow{9}{*}{$\norm{\delta\mbshat{\rho}}$ [$\mathrm{cm}$]}  & \multirow{3}{*}{EH} & Easy       & 1.706-\textcolor{Maroon}{2.479}-\textcolor{NavyBlue}{3.523}          & 1.694-\textcolor{Maroon}{2.537}-\textcolor{NavyBlue}{3.432}                            & 1.739-\textcolor{Maroon}{2.454}-\textcolor{NavyBlue}{3.401}                   & 1.572-\textcolor{Maroon}{\textbf{2.321}}-\textcolor{NavyBlue}{\textbf{3.357}}          & \textbf{1.563}-\textcolor{Maroon}{2.462}-\textcolor{NavyBlue}{3.460}                   \\
                                                                       &                     & Medium     & 1.984-\textcolor{Maroon}{3.327}-\textcolor{NavyBlue}{5.103}          & 2.074-\textcolor{Maroon}{3.323}-\textcolor{NavyBlue}{5.087}                            & 2.097-\textcolor{Maroon}{3.345}-\textcolor{NavyBlue}{5.056}                   & \textbf{1.881}-\textcolor{Maroon}{\textbf{3.252}}-\textcolor{NavyBlue}{\textbf{4.752}} & 1.940-\textcolor{Maroon}{3.340}-\textcolor{NavyBlue}{5.302}                            \\
                                                                       &                     & Hard       & 2.025-\textcolor{Maroon}{3.424}-\textcolor{NavyBlue}{4.448}          & 2.051-\textcolor{Maroon}{3.358}-\textcolor{NavyBlue}{4.311}                            & 2.072-\textcolor{Maroon}{3.406}-\textcolor{NavyBlue}{4.383}                   & 1.971-\textcolor{Maroon}{3.319}-\textcolor{NavyBlue}{\textbf{4.310}}                   & \textbf{1.949}-\textcolor{Maroon}{\textbf{3.316}}-\textcolor{NavyBlue}{4.373}          \\
        \cline{2-8}                                                    & \multirow{3}{*}{GZ} & Easy       & 1.511-\textcolor{Maroon}{2.105}-\textcolor{NavyBlue}{3.257}          & 1.572-\textcolor{Maroon}{2.002}-\textcolor{NavyBlue}{\textbf{3.136}}                   & 1.575-\textcolor{Maroon}{2.107}-\textcolor{NavyBlue}{3.244}                   & \textbf{1.498}-\textcolor{Maroon}{2.080}-\textcolor{NavyBlue}{3.338}                   & 1.505-\textcolor{Maroon}{\textbf{1.920}}-\textcolor{NavyBlue}{3.187}                   \\
                                                                       &                     & Medium     & 1.726-\textcolor{Maroon}{2.426}-\textcolor{NavyBlue}{3.213}          & 1.576-\textcolor{Maroon}{2.312}-\textcolor{NavyBlue}{2.926}                            & 1.523-\textcolor{Maroon}{1.952}-\textcolor{NavyBlue}{2.683}                   & \textbf{1.471}-\textcolor{Maroon}{\textbf{1.951}}-\textcolor{NavyBlue}{\textbf{2.594}} & 1.592-\textcolor{Maroon}{2.103}-\textcolor{NavyBlue}{3.160}                            \\
                                                                       &                     & Hard       & 1.461-\textcolor{Maroon}{3.619}-\textcolor{NavyBlue}{19.29}          & 1.261-\textcolor{Maroon}{1.902}-\textcolor{NavyBlue}{3.810}                            & \textbf{1.191}-\textcolor{Maroon}{1.656}-\textcolor{NavyBlue}{\textbf{2.768}} & 1.206-\textcolor{Maroon}{\textbf{1.644}}-\textcolor{NavyBlue}{5.585}                   & 1.255-\textcolor{Maroon}{1.714}-\textcolor{NavyBlue}{2.794}                            \\
        \cline{2-8}                                                    & \multirow{3}{*}{MP} & Easy       & 2.725-\textcolor{Maroon}{3.419}-\textcolor{NavyBlue}{4.374}          & \textbf{2.393}-\textcolor{Maroon}{3.176}-\textcolor{NavyBlue}{4.336}                   & 2.438-\textcolor{Maroon}{3.071}-\textcolor{NavyBlue}{\textbf{4.052}}          & 2.402-\textcolor{Maroon}{\textbf{2.825}}-\textcolor{NavyBlue}{4.188}                   & 2.567-\textcolor{Maroon}{3.867}-\textcolor{NavyBlue}{4.146}                            \\
                                                                       &                     & Medium     & 3.270-\textcolor{Maroon}{5.621}-\textcolor{NavyBlue}{7.452}          & 3.243-\textcolor{Maroon}{5.703}-\textcolor{NavyBlue}{7.027}                            & \textbf{2.768}-\textcolor{Maroon}{\textbf{5.093}}-\textcolor{NavyBlue}{6.688} & 2.892-\textcolor{Maroon}{5.096}-\textcolor{NavyBlue}{\textbf{6.615}}                   & 3.228-\textcolor{Maroon}{5.509}-\textcolor{NavyBlue}{6.796}                            \\
                                                                       &                     & Hard       & 4.059-\textcolor{Maroon}{7.621}-\textcolor{NavyBlue}{23.76}          & \textbf{3.472}-\textcolor{Maroon}{\textbf{6.093}}-\textcolor{NavyBlue}{\textbf{9.565}} & 3.495-\textcolor{Maroon}{6.115}-\textcolor{NavyBlue}{11.43}                   & 3.655-\textcolor{Maroon}{6.362}-\textcolor{NavyBlue}{33.99}                            & 3.670-\textcolor{Maroon}{6.490}-\textcolor{NavyBlue}{10.74}                            \\ \hline
        Iterations                                                     & ALL                 & ALL        & 7                                                                   & 7                                                                                     & \textbf{6}                                                                   & \textbf{6}                                                                            & 7                                                                                     \\
        Time [$\mathrm{s}$]                                            & ALL                 & ALL        & 6.33                                                                & \textbf{1.24}                                                                         & 4.23                                                                         & 4.25                                                                                  & 4.40                                                                                  \\
        \bottomrule
    \end{tabular}
\end{table*}
The proposed methods are compared against GNC-TLS and GNC-GM proposed by Yang \etal in \cite{Yang2020GNC} in terms of the alignment errors and processing time.
Table~\ref{tab:icp_results} shows results for each dataset and difficulty combinations.
The errors are reported as 50\%-\textcolor{Maroon}{75\%}-\textcolor{NavyBlue}{90\%} percentiles to show not only the accuracy of the results but also the variance.

Overall, all GNC loss functions perform well in the structured environments (EH) regardless of the initial perturbation, with GNC-ADAPT performing marginally better than the others.
However, the difference in performance becomes more pronounced in semistructured (GZ) and unstructured (MP) environments.
Semistructured environments exhibit higher median errors in rotation than structured environments, whereas unstructured environments exhibit higher median errors in both rotation and translation.

The difference between the GNC-GM by Yang \etal \cite{Yang2020GNC} and the proposed GNC-GM is from its derivation.
The former tailors GNC specifically to GM, and the latter tailors GNC using the proposed method with a fixed shape parameter ${\alpha}$ in \eqref{eq:agnc_sf3}.
Even though GNC is applied to the same robust loss function, the proposed GNC-GM shows a slight improvement in the median and variance of the errors especially with the ``MP'' dataset at a cost of processing time.
The comparison here between these two methods is to show that the GNC-enabled adaptive function performs on par or better than the state-of-the-art GNC approach applied to a specific loss function.
Further, the use of an adaptive function removes from the user the choice of a dedicated loss function.

The main advantage of GNC-ADAPT is that the user does not need to specify the shape parameter ${\alpha}$ as it is automatically determined from the data.
Although in some cases like the ``GZ'' dataset with ``hard'' difficulty, GNC-GM performs better than GNC-ADAPT, the difference is marginal.
This would happen when the shape parameter ${\alpha}$ computed via optimization is close to the shape of the traditional loss function, like GM in this case, but ${\alpha}$ is not guaranteed to be optimal.
The effect of a non-optimal shape parameter is highlighted in the variance of the errors, especially in the ``MP'' dataset with ``hard'' difficulty.
Regardless, GNC-ADAPT outperforms in most cases, even in comparison to GNC-TLS, highlighting the robustness of GNC-ADAPT to variations in environment and initial perturbation.

While GNC-AMB performs well across environments and difficulty levels, its robustness is notable for difficult initial perturbations in unstructured environments.
For example, in the experiment with the ``MP'' dataset with the ``hard'' difficulty level, 90\% of the rotation error for GNC-ADAPT falls below ${\SI{8.597}{\degree}}$, whereas GNC-AMB performs with ${\SI{0.896}{\degree}}$.
The same trend is observed in the translation error, where the 90th percentile of the error is ${\SI{33.99}{\centi\metre}}$ for GNC-ADAPT and ${\SI{10.74}{\centi\metre}}$ for GNC-AMB.
The reduction in variance shows that GNC-AMB is more robust to various types of problems than GNC-ADAPT.

\subsubsection{Processing Time}
\begin{figure}[tb]
    \centering
    \includegraphics[width=\linewidth]{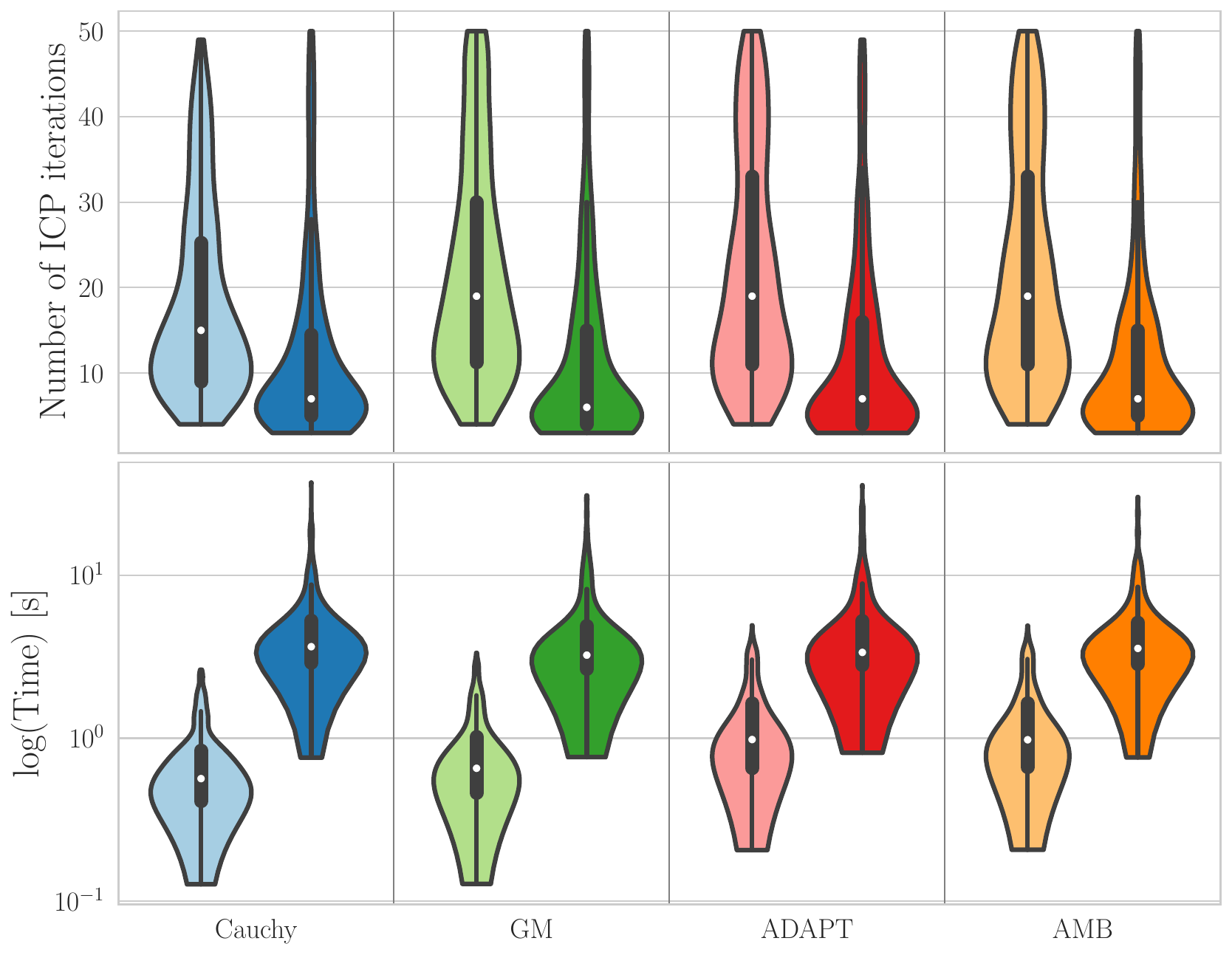}
    \caption{
        Violin plots show the number of iterations and convergence time of the converged ICP trials.
        The times are shown in a log scale for clear representation.
        The proposed GNC-enabled methods and their corresponding non-GNC baseline functions are shown in the same color with a higher and lower opacity, respectively.
        The width of each violin distribution is proportional to the number of successful trials.
    }
    \label{fig:convergence_time}
\end{figure}

Figure~\ref{fig:convergence_time} shows the number of iterations and log-scaled convergence time of the converged ICP trials.
Note that this study was conducted using non-optimized \texttt{MATLAB} code, and the timing results are included for relative comparison.

Overall, GNC-incorporated methods take fewer iterations but more time to converge than their non-GNC counterparts, as the inner GNC loop from Algorithm~\ref{alg:gnc-alf} must be run whenever new point correspondences are found.
GNC-incorporated methods require a significantly longer time to converge, which may hinder their practicality for real-time applications.
However, the proposed methods exhibit lower processing time than GNC-TLS with more robustness and a larger reduction in errors.

\subsection{Mesh Registration}
\label{sec:results_mr}

Mesh registration is an algorithm used to find a 3D rigid transformation ${\mbf{T}^{c_{m}c_{p}}_{mp} \in SE(3)}$ between a 3D mesh and a point cloud, where ${c_{m}}$ and ${c_{p}}$ are the centroids of mesh and point cloud, respectively.
A mesh ${\{\{\mbf{r}^{q_{ij}c_{m}}_{m}\}_{j=1}^{M_{i}}, \mbf{n}^{m_{i}}_{m}\}_{i=1}^{N_{s}}}$ is represented as a collection of ${N}_{s}$ surfaces with corresponding unit normal vector ${\mbf{n}^{m_{i}}_{m}}$ resolved in the mesh frame ${\mc{F}_{m}}$.
Each surface is represented as a set of vertices ${\{\mbf{r}^{q_{ij}c_{m}}_{m}\}_{j=1}^{M_{i}}}$ where ${M_{i}}$ is the number of vertices for ${i}$-th surface, and ${q_{ij}}$ is the ${j}$-th vertex of ${i}$-th surface.
A point cloud is ${\{\mbf{r}^{p_{i}c_{p}}_{p}, \mbf{n}^{p_{i}}_{p}\}_{i=1}^{N_{p}}}$ is represented as a collection of ${N_{p}}$ points with estimated normals ${\mbf{n}^{p_{i}}_{p}}$ resolved in the point-cloud frame ${\mc{F}_{p}}$.
For the brevity of notation, ${\mbf{T}_{mp} = \mbf{T}^{c_{m}c_{p}}_{mp}}$, ${\mbf{r}^{pm}_{m} = \mbf{r}^{c_{p}c_{m}}_{m}}$, ${\mbf{r}^{i}_{p} = \mbf{r}^{p_{i}c_{p}}_{p}}$, ${\mbf{n}^{i}_{m} = \mbf{n}^{m_{i}}_{m}}$, and ${\mbf{n}^{i}_{p} = \mbf{n}^{p_{i}}_{p}}$.

The least-squares formulation for the mesh registration problem is based on \cite{Yang2020GNC}.
Given a set of ${N}$ putative correspondences from a point cloud to a mesh, mesh registration finds the rigid body transformation that best aligns the point cloud to the mesh by solving
\begin{align}
    \label{eq:mr}
    \mbf{T}^{\star}_{mp} & = \underset{\mbf{T}_{mp} \in SE(3)}{\argmin} \sum_{i=1}^{N} \onehalf w_{i}\left(\f{1}{\sigma^{pi}_{p}}{e^{i}_{p}}^{2} + \norm{\mbf{e}^{i}_{n}}_{{\mbs{\Sigma}^{i}_{n}}^{-1}}\right),
\end{align}
where ${\sigma_{p}^{pi}}$ and ${\mbs{\Sigma}^{i}_{n}}$ are the covariance associated with the point-to-plane and normal-to-normal distance, given respectively as
\begin{align}
    e^{i}_{p}\left(\mbf{T}_{mp}, \mbftilde{r}^{i}_{m}, \mbf{r}^{i}_{p}, \mbf{n}^{i}_{m}\right) & = {\mbf{n}^{i}_{m}}^{\trans}\left(\mbf{C}_{mp}\mbf{r}^{i}_{p} + \mbf{r}^{pm}_{m} - \mbftilde{r}^{i}_{m}\right), \\
    \mbf{e}^{i}_{n}\left(\mbf{T}_{mp}, \mbf{n}^{i}_{m}, \mbf{n}^{i}_{p}\right)                 & = \mbf{n}^{i}_{m} - \mbf{C}_{mp}\mbf{n}^{i}_{p}.
\end{align}
For each point-to-face association, an arbitrary point ${\mbftilde{r}^{i}_{m}}$ is chosen from the corresponding surface.

Unlike PCA, the mesh registration experiment is performed on sparse point-to-face correspondences to test the robustness of the loss functions.
The shape function from Example~\ref{ex:3} was used to generate results for the proposed methods in this section.
\begin{figure}[tb]
    \centering
    \includegraphics[width=\linewidth]{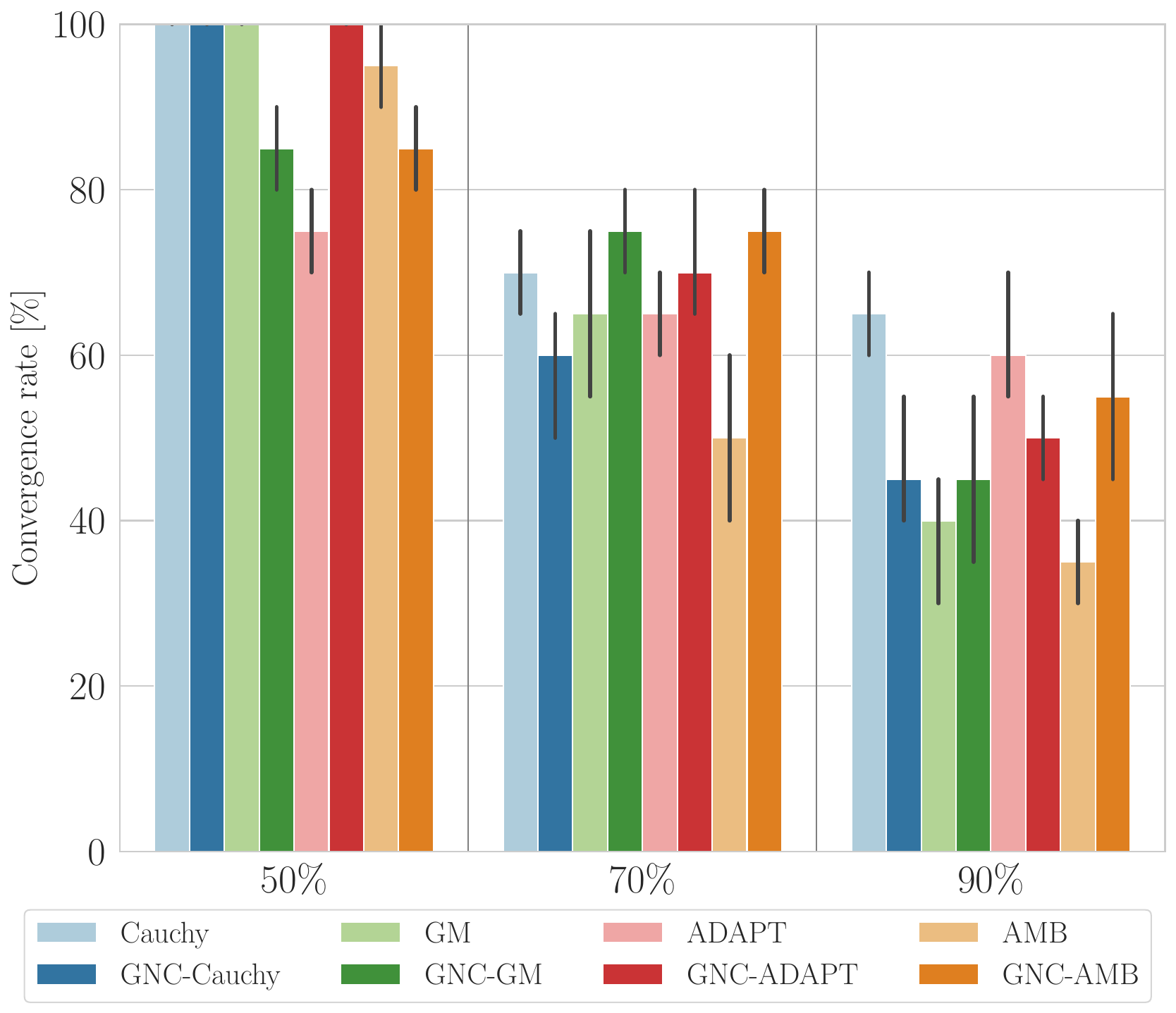}
    \caption{
        Mesh registration convergence for every baseline robust loss function (transparent) and their GNC variants (opaque) are shown for different outlier ratios.
        The solver is considered converged when the relative change in the cost is less than \SI{1e-8}{}.
    }
    \label{fig:mr_sim_convergence}
\end{figure}
This experiment evaluates the convergence rate and alignment errors in the presence of various outlier ratios.

\subsubsection{Simulation Setting}
The simulation setup is based on \cite{Yang2023Certifiably}.
A set of ${N=20}$ 3D planes ${\{\mbf{r}^{i}_{m}, \mbf{n}^{i}_{m}\}_{i=1}^{N}}$ are randomly generated, where the points are sampled from ${\mbf{r}^{i}_{m} \sim \mc{N}(\mbf{0}, \mbf{1})}$.
Then, a random point ${\mbftilde{r}^{i}_{m}}$ on the same plane as ${\mbf{r}^{i}_{m}}$ is generated via ${\mbftilde{r}^{i}_{m} = \mbf{r}^{i}_{m} + {\mbf{w}^{i}_{m}}^{\times}\mbf{n}^{i}_{m}}$, where ${\mbf{w}^{i}_{m} \sim \mc{N}(\mbf{0}, \mbf{1})}$.
Using an arbitrary rigid body transformation ${\mbf{T}_{mp}}$, a corresponding point cloud ${\{\mbf{r}^{i}_{p}, \mbf{n}^{i}_{p}\}}$ is generated by
\begin{align}
    \bbm {\mbf{r}^{i}_{p}}                                                                                                                                     \\ 1 \ebm & = \mbf{T}_{mp}^{-1} \bbm {\mbf{r}^{i}_{m}} \\ 1 \ebm + \bbm {\mbs{\sigma}^{pi}_{p}} \\ 0 \ebm, \\
    \mbf{n}^{i}_{p} & = \f{\mbf{C}_{mp}^{\trans}\mbf{n}^{i}_{m} + \mbs{\sigma}^{ni}_{p}}{\norm{\mbf{C}_{mp}^{\trans}\mbf{n}^{i}_{m} + \mbs{\sigma}^{ni}_{p}}},
\end{align}
where ${\mbs{\sigma}^{pi}_{p}, \mbs{\sigma}^{ni}_{p} \sim \mc{N}(\mbf{0}, 0.01^{2}\mbf{1})}$.
Note that both generated mesh and point cloud are unitless.
Given the mesh and the noisy point cloud, \eqref{eq:mr} is solved using the \texttt{mosek} SDP solver, as was done in \cite{Yang2023Certifiably}.
No initial transformation between the point cloud and the mesh is provided to the solver.

The performance of the solver is evaluated using the estimation error, ${\log(\delta\mbf{T}) = \log({\mbf{T}_{mp}^{-1}\mbf{T}^{\star}_{mp}}) \in \mathfrak{se}(3)}$, where ${\mbf{T}_{mp}}$ is the ground-truth transformation, and the processing time.
For simulation experiments, statistics are computed over 20 Monte Carlo runs per setup as done in \cite{Yang2020GNC}.
Each setup was performed over various outlier ratios that increased from 50\% to 90\% in 20\% increments.

The solver is considered to have converged when the difference in cost between two consecutive iterations is less than ${\SI{1e-8}{}}$ before the maximum number of iterations is reached.
The maximum number of iterations is set to 50, and the solver is considered to have failed if the maximum number of iterations is reached without convergence.

\subsubsection{Simulation Results}
\label{sec:mr_sim_results}
\begin{table*}[tb]
    \renewcommand{\arraystretch}{1.25} 
    \centering
    \caption{
        Simulated mesh registration results comparing GNC-TLS \cite{Yang2020GNC} to GNC-AMB and GNC-ADAPT.
        Note GNC-GM (from proposed) is simply GNC-ADAPT with ${\alpha^{\star} = -2}$.
        Error percentiles, mean processing times, and median numbers of iterations to converge are shown for different outlier ratios and compared with those of GNC-TLS.
        The errors are reported as 50\%-\textcolor{Maroon}{75\%}-\textcolor{NavyBlue}{90\%} percentiles.
        The lowest error in each percentile as well as the fewest iterations in each row are highlighted in bold.
    }
    \label{tab:mr_sim_results}
    \begin{tabular}{l|c|cccccc}
        \toprule
        Metric                                                         & Outlier & GNC-TLS                                                                      & GNC-GM \cite{Yang2020GNC}                                                             & GNC-GM                                                                                & GNC-ADAPT                                                                             & GNC-AMB                                                                      \\ \hline
        \multirow{3}{*}{$\norm{\delta\mbshat{\phi}}$ [$\mathrm{deg}$]} & 50\%    & \textbf{0.246}-\textcolor{Maroon}{\textbf{0.291}}-\textcolor{NavyBlue}{9.323} & 0.267-\textcolor{Maroon}{0.353}-\textcolor{NavyBlue}{8.667}                            & 0.267-\textcolor{Maroon}{0.353}-\textcolor{NavyBlue}{8.667}                            & 0.368-\textcolor{Maroon}{0.685}-\textcolor{NavyBlue}{9.598}                            & 1.872-\textcolor{Maroon}{5.169}-\textcolor{NavyBlue}{\textbf{8.228}}          \\
                                                                       & 70\%    & 74.55-\textcolor{Maroon}{146.1}-\textcolor{NavyBlue}{150.8}                   & 39.35-\textcolor{Maroon}{125.3}-\textcolor{NavyBlue}{164.8}                            & 39.35-\textcolor{Maroon}{123.8}-\textcolor{NavyBlue}{164.8}                            & \textbf{36.40}-\textcolor{Maroon}{123.8}-\textcolor{NavyBlue}{151.1}                   & 57.25-\textcolor{Maroon}{\textbf{111.2}}-\textcolor{NavyBlue}{\textbf{130.6}} \\
                                                                       & 90\%    & 139.3-\textcolor{Maroon}{156.6}-\textcolor{NavyBlue}{162.2}                   & 143.1-\textcolor{Maroon}{157.9}-\textcolor{NavyBlue}{175.0}                            & 137.5-\textcolor{Maroon}{154.7}-\textcolor{NavyBlue}{163.0}                            & \textbf{136.6}-\textcolor{Maroon}{\textbf{153.8}}-\textcolor{NavyBlue}{\textbf{162.9}} & 139.3-\textcolor{Maroon}{155.2}-\textcolor{NavyBlue}{169.1}                   \\ \hline
        \multirow{3}{*}{$\norm{\delta\mbshat{\rho}}$}                  & 50\%    & 0.010-\textcolor{Maroon}{0.014}-\textcolor{NavyBlue}{0.109}                   & \textbf{0.009}-\textcolor{Maroon}{\textbf{0.017}}-\textcolor{NavyBlue}{\textbf{0.139}} & \textbf{0.009}-\textcolor{Maroon}{\textbf{0.017}}-\textcolor{NavyBlue}{\textbf{0.139}} & 0.014-\textcolor{Maroon}{0.033}-\textcolor{NavyBlue}{0.181}                            & 0.077-\textcolor{Maroon}{0.155}-\textcolor{NavyBlue}{0.428}                   \\
                                                                       & 70\%    & 0.641-\textcolor{Maroon}{1.552}-\textcolor{NavyBlue}{2.654}                   & \textbf{0.411}-\textcolor{Maroon}{1.357}-\textcolor{NavyBlue}{2.350}                   & \textbf{0.411}-\textcolor{Maroon}{1.321}-\textcolor{NavyBlue}{1.705}                   & 0.534-\textcolor{Maroon}{1.402}-\textcolor{NavyBlue}{1.702}                            & 0.818-\textcolor{Maroon}{\textbf{1.223}}-\textcolor{NavyBlue}{\textbf{1.623}} \\
                                                                       & 90\%    & 1.703-\textcolor{Maroon}{3.304}-\textcolor{NavyBlue}{4.718}                   & \textbf{1.119}-\textcolor{Maroon}{2.345}-\textcolor{NavyBlue}{3.876}                   & 1.382-\textcolor{Maroon}{2.797}-\textcolor{NavyBlue}{3.876}                            & 1.197-\textcolor{Maroon}{2.642}-\textcolor{NavyBlue}{3.711}                            & 1.159-\textcolor{Maroon}{\textbf{1.382}}-\textcolor{NavyBlue}{\textbf{1.853}} \\ \hline
        Iterations                                                     & ALL     & \textbf{7}                                                                   & 28                                                                                    & 29                                                                                    & 28                                                                                    & 28                                                                           \\
        \bottomrule
    \end{tabular}
\end{table*}
\begin{figure*}
    \centering
    \begin{subfigure}[t]{0.49\linewidth}
        \centering
        \includegraphics[width=\linewidth]{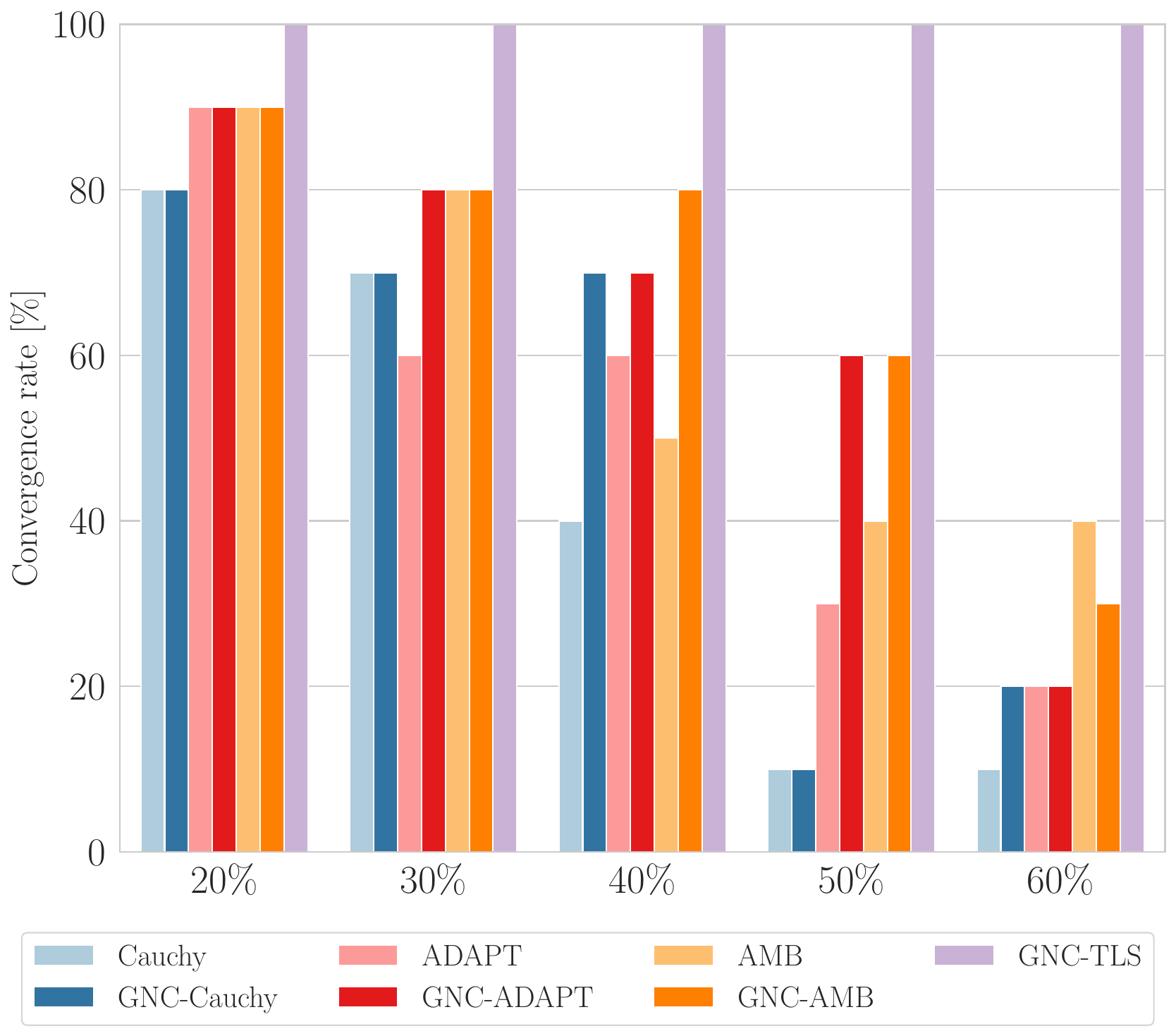}
        \caption{Percentage of converged trials.}
        \label{fig:mr_real_result_convergence}
    \end{subfigure}
    \begin{subfigure}[t]{0.49\linewidth}
        \centering
        \includegraphics[width=\linewidth]{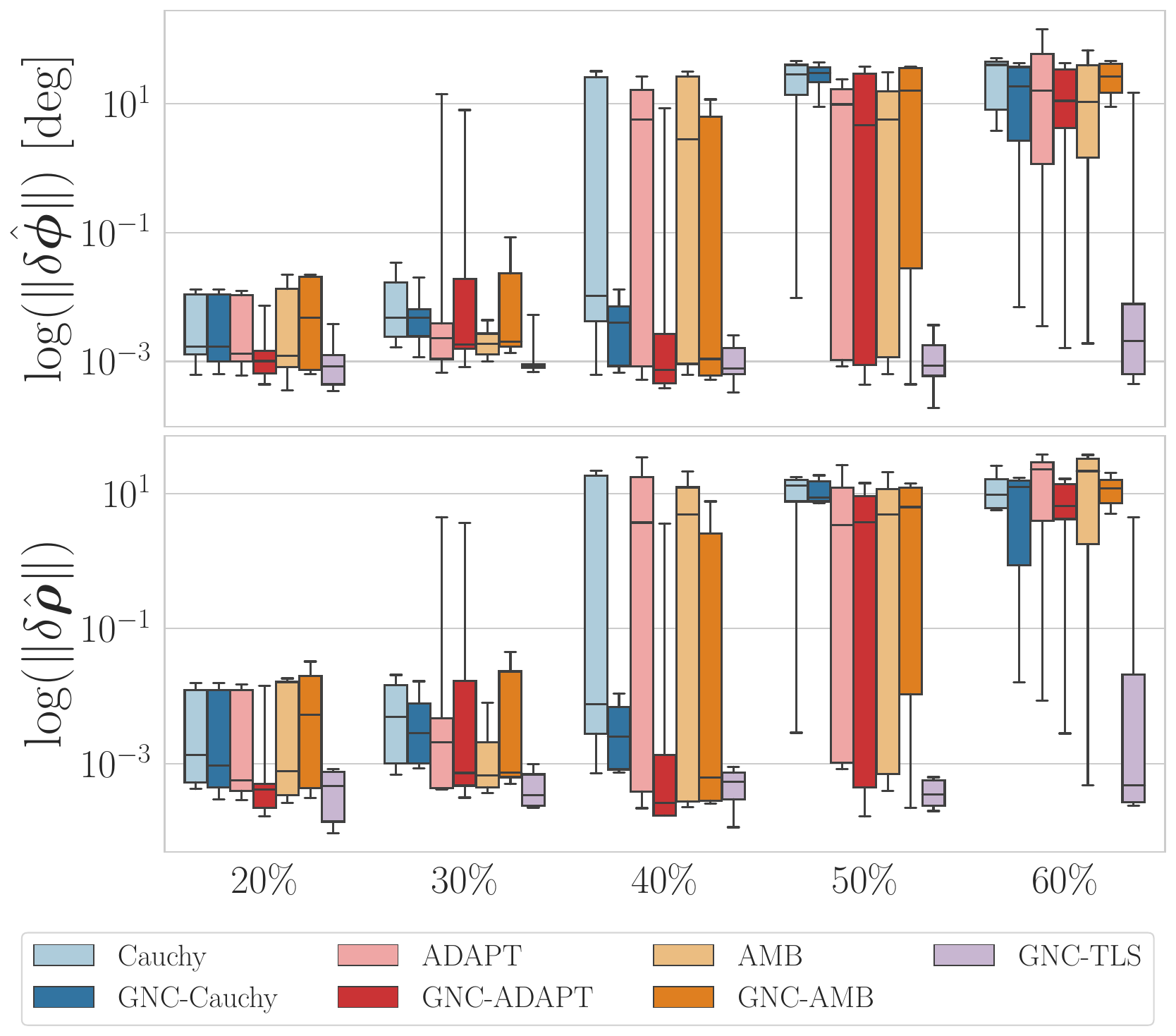}
        \caption{Rotation and translation error.}
        \label{fig:mr_real_result_error}
    \end{subfigure}
    \caption{
        Mesh registration results on the \texttt{HomebrewedDB} dataset.
        The left figure shows the percentage of converged trials for different outlier ratios.
        The right figure shows the median and variance of the errors for different outlier ratios.
    }
    \label{fig:mr_real_result}
\end{figure*}
The percentage of converged trials of the loss functions for different outlier ratios is shown in Figure~\ref{fig:mr_sim_convergence}.
The adaptive methods, ADAPT and AMB, show a lower convergence rate than the non-adaptive methods, Cauchy and GM, for most outlier ratios.
The reduction in convergence rate is due to the lack of sufficient statistics to estimate the shape parameter ${\alpha}$.
At a 50\% outlier ratio, Cauchy and GM converge in all runs, while ADAPT and AMB converge in 75\% and 95\% of the runs, respectively.

However, in many cases, the proposed GNC-enabled loss functions have a larger percentage of converged trials compared to their non-GNC baseline.
For example, GNC with GM converges 10\% more at the 70\% outlier level and 5\% more at the 90\% outlier level than the baseline GM.
The increase in convergence rate is more pronounced in AMB, where the GNC variant shows a 25\% and 20\% improvement in convergence rate at the 70\% and 90\% outlier ratios, respectively.
Simulation results show that there is an increase in convergence rate for the GNC variants especially for the adaptive loss functions.

The registration errors of the proposed methods are compared against the state-of-the-art GNC loss functions in Table~\ref{tab:mr_sim_results}.
Note that GNC-GM with the proposed method performs very similarly to GNC-GM by \cite{Yang2020GNC} in terms of the median error and variance.
This ensures that the proposed method with the fixed shape parameter ${\alpha}$ is as effective as manually tailoring GNC to a specific loss function.

All methods perform poorly in the presence of a large number of outliers.
In terms of the median translation error, GNC-GM performs the best in all outlier ratios followed by GNC-ADAPT.
However, GNC-AMB exhibits less variance in errors by outperforming the other methods in the 90th percentile of the translation error with \SI{1.623}{} and \SI{1.853}{} at 70\% and 90\% outlier ratio, respectively.

\subsubsection{Experiment Setting}
\begin{figure}[tb]
    \centering
    \begin{subfigure}[t]{0.49\linewidth}
        \centering
        \includegraphics[width=\linewidth]{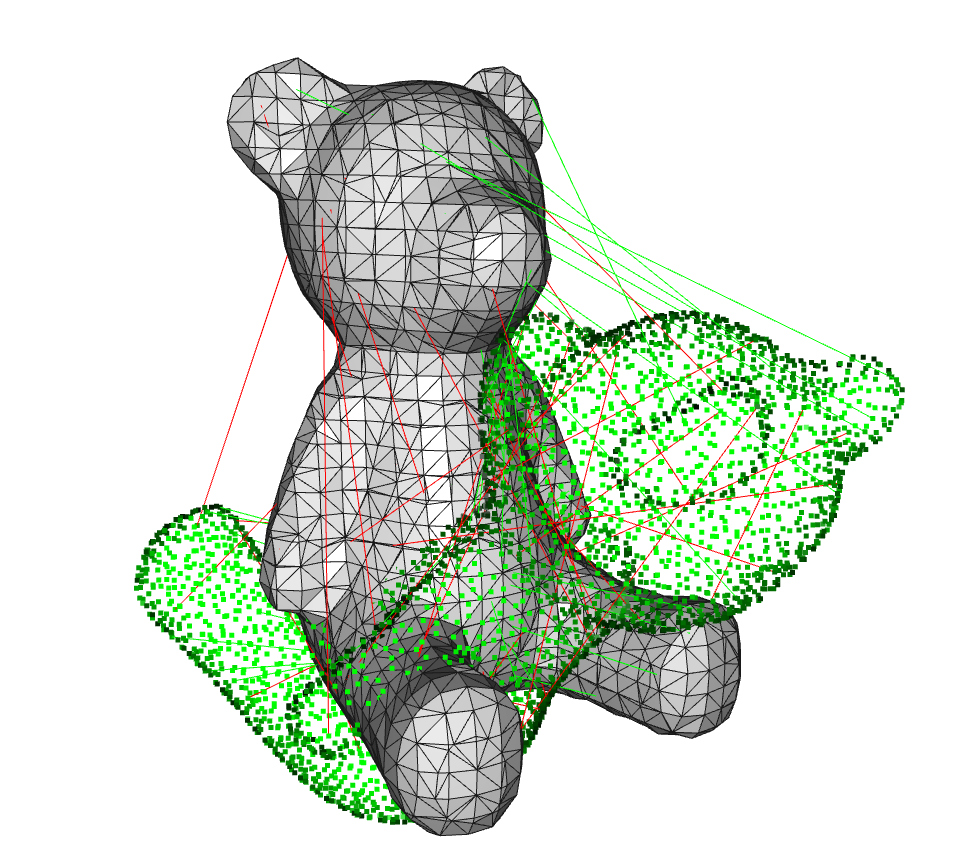}
        \caption{Prior}
        \label{fig:mr_problem_prior}
    \end{subfigure}
    \begin{subfigure}[t]{0.49\linewidth}
        \centering
        \includegraphics[width=\linewidth]{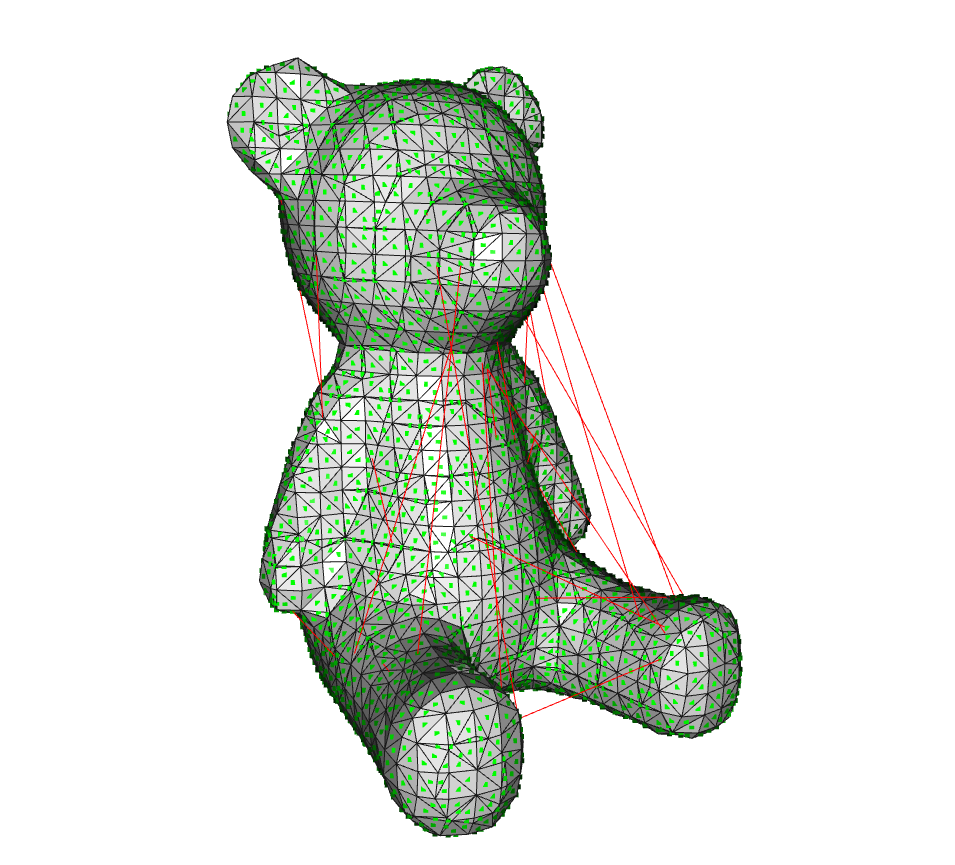}
        \caption{Posterior}
        \label{fig:mr_problem_posterior}
    \end{subfigure}
    \caption{
        An illustration of the mesh registration experiment with a 40\% outlier ratio is shown here.
        Green lines are the inlier point-to-face correspondences, and red lines are the outliers.
        The right figure shows the registered mesh to point cloud using the proposed method GNC-AGNC.
    }
    \label{fig:mr_problem}
\end{figure}
The \texttt{HomebrewedDB} dataset \cite{Kaskman2019HomebrewedDB} is used to evaluate the performance of the proposed robust loss functions on real-world data.
Among the 33 objects featured by the dataset, ``Teddy Bear'' mesh is used.

In each Monte Carlo trial, a noisy point cloud is generated by sampling points from the center of each surface of the mesh and adding Gaussian noise with standard deviation ${\sigma=0.01}$.
The sampled points that are in the mesh frame are transformed to the point-cloud frame with randomly chosen ground-truth rigid-body transformation ${\mbf{T}_{mp} \in SE(3)}$
Using the transformed point cloud, surface normals are estimated using the \texttt{Open3D} library \cite{Zhou2018Open3D}.
Then, ${N=100}$ point-to-face correspondences are randomly selected and amongst the ${100}$ correspondences, some are modified as outliers with 10\% to 70\% ratio with a 10\% increment as shown in Figure~\ref{fig:mr_problem_prior}.
The mesh registration problem is solved using the same setup as the simulation experiment over ${10}$ Monte Carlo trials.

\subsubsection{Experiment Results}
The percentage of converged trials and the corresponding rotation and translation errors for different outlier ratios are displayed in Figure~\ref{fig:mr_real_result}.
The non-GNC functions are shown in transparent colors, and the proposed GNC-enabled functions are shown in opaque.
The state-of-the-art GNC-TLS is shown in yellow.

Although GNC-TLS outperforms the proposed methods at all outlier ratios in terms of convergence rate and errors, the results show the improvement of the baseline functions after being tailored to GNC.
For example at 40\% and 50\% outlier ratio, the Cauchy, ADAPT, and AMB functions have shown improvement in convergence rate with GNC.
Also, the baseline functions start to fail from a 40\% outlier ratio with large errors in rotation and translation as shown in Figure~\ref{fig:mr_real_result_error}, but the GNC-incorporates still manage to converge successfully.
As discussed in Section~\ref{sec:mr_sim_results}, the performance of the adaptive methods is worse than the non-adaptive methods due to the lack of sufficient statistics to estimate the shape parameter ${\alpha}$.

\subsection{Pose-graph Optimization}
\label{sec:results_pgo}

Pose-graph optimization (PGO) is a problem of estimating the poses of a robot given a set of noisy relative pose measurements.
Recently, robust functions have been used extensively to solve PGO problems \cite{Olson2008Robust, Agarwal2013Robust, McGann2023Robust}.
The proposed methods are therefore tested and evaluated on a set of PGO problems to show their effectiveness and robustness against state-of-the-art robust loss functions in the presence of erroneous loop-closures that occur due to data association errors caused by, for example, perceptual aliasing.

The PGO problem is formulated as a least-squares problem, where the objective function is to minimize the sum of the squared residuals of the relative pose measurements.
The objective function is given as
\begin{align}
    \label{eq:pgo}
    \mbf{X}^{\star} & = \underset{\mbf{X}}{\argmin} \sum_{(i, j) \in \mc{E}} \rho\left(\mbf{e}^{ij}\right),
\end{align}
where ${\mbf{X} = \{\mbf{X}_{1}, \ldots, \mbf{X}_{N}\}}$ is the set of poses, ${\mc{E}}$ is the set of relative pose measurements, and ${\mbf{e}^{ij}}$ is the residual of the relative pose measurement between ${\mbf{X}_{i}}$ and ${\mbf{X}_{j}}$.
The residual ${\mbf{e}^{ij}}$ is given as
\begin{align}
    \mbf{e}^{ij} & = \log\left(\mbf{X}_{i}^{-1}\mbf{X}_{j}\mbf{X}_{ij}^{-1}\right)^{\vee},
\end{align}
where ${\mbf{X}_{ij}}$ is the relative pose measurement between ${\mbf{X}_{i}}$ and ${\mbf{X}_{j}}$.
The residual ${\mbf{e}^{ij}}$ is a vector representing the difference between the relative pose measurement and the actual relative pose.

\subsubsection{Experiment Setting}
\begin{figure}[tb]
    \centering
    \begin{subfigure}[t]{0.49\linewidth}
        \centering
        \includegraphics[width=\linewidth]{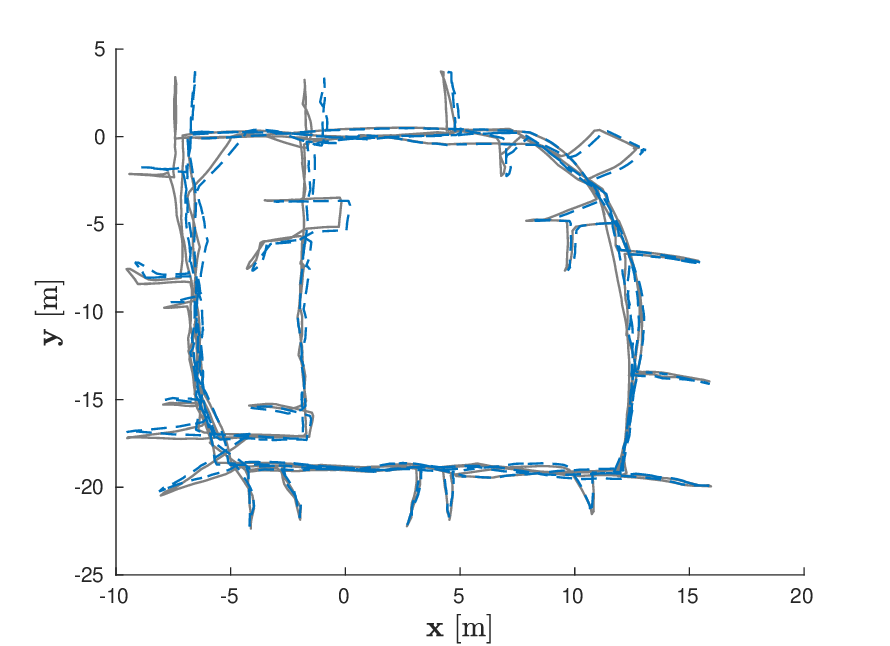}
        \caption{Low noise $\sigma^{\theta} = 0.001$}
        \label{fig:pgo_problem_small_noise}
    \end{subfigure}
    \begin{subfigure}[t]{0.49\linewidth}
        \centering
        \includegraphics[width=\linewidth]{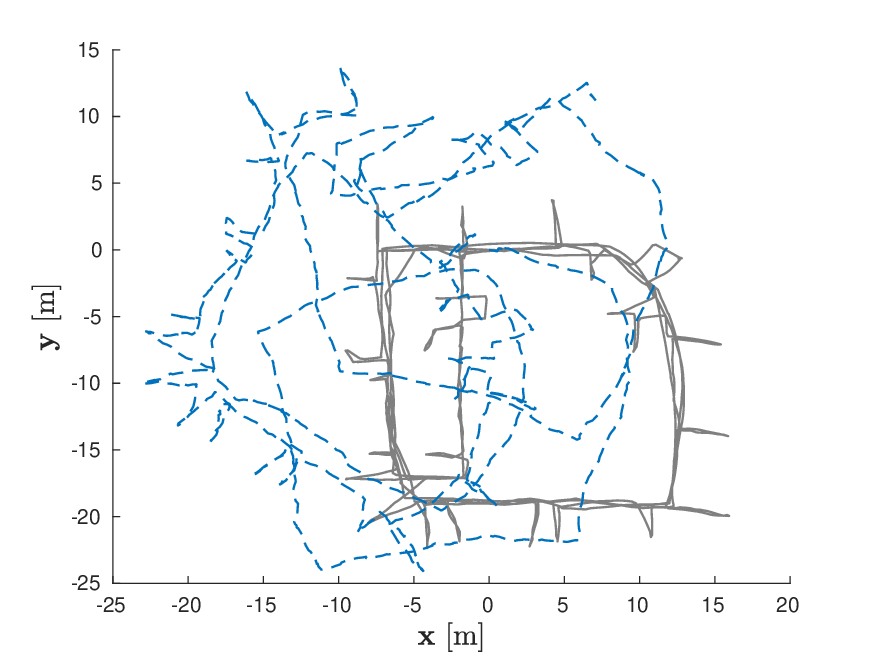}
        \caption{High noise $\sigma^{\theta} = 0.01$}
        \label{fig:pgo_problem_large_noise}
    \end{subfigure}
    \caption{
        Example Intel dataset with varying odometry noise.
        The trajectories' ground-truth from \cite{Rosen2019SESync} is shown in gray, while the dead-reckoning trajectory is shown in blue.
    }
    \label{fig:pgo_problem}
\end{figure}
The PGO problem is formulated using the \texttt{g2o} library \cite{Kummerle2011g2o}.
To provide continuity with prior works, MIT CSAIL \cite{Carlone2014Fast}, Manhattan \cite{Olson2006Fast}, and the Intel Research Lab Dataset \cite{Carlone2014From} have been used.

A key aspect of the proposed method is its robustness to poor initialization.
In practice, as the odometry noise increases, the states in the optimization problem can be initialized far from their ground-truth values.
Therefore, by varying the amount of odometry noise, the robustness of the proposed method to poor initialization is evaluated.
The effects of odometry noise are shown in Figure~\ref{fig:pgo_problem}.

Another aspect of the proposed method is the ability to handle outliers.
The robustness of the algorithm is evaluated over different quantities of outlier loop-closure measurements with a high odometry noise value.
Outliers are generated by pairing up poses and adding a random rotation and translation to the relative pose measurement.
The outlier ratio from 10\% to 50\% is tested in 10\% increments with the odometry noise ${\sigma^{\theta} = \SI{0.01}{\degree}}$.

The solvers are considered to have converged when the relative change in the cost is less than ${\SI{1e-6}{}}$ before the maximum number of iterations, ${50}$, is reached.
The effectiveness of the solvers on the PGO problem is evaluated using the absolute trajectory error (ATE) as it is widely used to evaluate SLAM algorithms,
\begin{align}
    \textrm{ATE} = \sqrt{\f{1}{K} \sum_{k=1}^{K}\norm{\log({\mbf{X}_{k}^{-1}\mbf{X}_{k}^{\star}})^{\vee}}},
\end{align}
where ${K}$ is the number of poses in a problem, and ${\mbf{X}^{\star}}$ is the ground-truth pose obtained via SE-Sync \cite{Rosen2019SESync} using only inlier measurements.
The proposed methods are compared against SE-Sync as it is capable of recovering certifiably globally optimal solutions.
Ten Monte Carlo trials are performed for each odometry noise level and outlier ratio, and the absolute trajectory error (ATE) is evaluated for each trial.

\subsubsection{Experiment Results}
\begin{figure*}[tb]
    \centering
    \includegraphics[width=\linewidth]{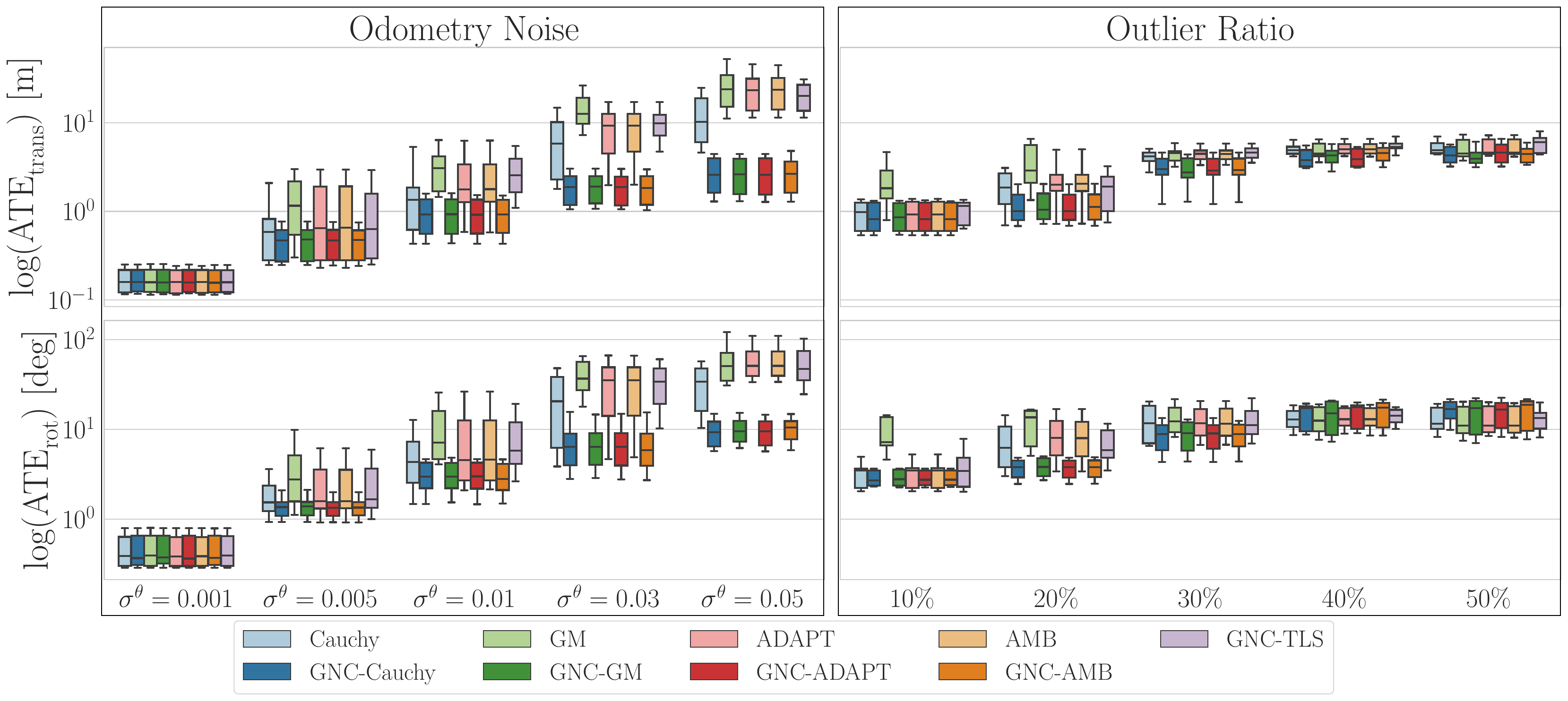}
    \caption{
        The absolute trajectory errors for each noise level (left) and outlier ratio (right) are shown in a log scale.
        The non-GNC robust loss functions are shown in transparent colors, and the proposed GNC-enabled functions are shown in opaque with the state-of-the-art GNC-TLS in purple.
    }
    \label{fig:pgo_results}
\end{figure*}
The proposed methods are evaluated for the robustness to poor initialization and the robustness to outliers.
In the left figure of Figure~\ref{fig:pgo_results}, the ATEs for each noise level are shown in a log scale.
The non-GNC robust loss functions are shown in transparent colors, and the proposed GNC-enabled functions are shown in opaque with the state-of-the-art GNC-TLS in purple.
The results show that the GNC variants are more robust to poor initialization and the effect is more pronounced as the odometry noise increases.
When the optimization problem is initialized with a large amount of odometry noise, the non-GNC loss functions consider the true loop-closure measurements as outliers and converge to a suboptimal solution.
However, the proposed GNC-enabled functions are able to recover the ground-truth trajectory with high precision.
Also, the proposed methods outperform the state-of-the-art GNC-TLS in different initialization conditions.

The right figure of Figure~\ref{fig:pgo_results} shows the percentage of converged trials for different outlier ratios.
The proposed GNC-enabled functions show a higher percentage of converged trials than the non-GNC loss functions.
Especially, the GNC-incorporated functions based on the proposed method outperform GNC-TLS in both rotation and translation ATE.
However, as the outlier ratio increases, the percentage of converged trials decreases for all loss functions.
In a high outlier and high noise scenario, the first few iterations of the optimization problem with a GNC-incorporated loss function are dominated by the outliers, and the solver is unable to achieve the optimal trajectory.
However, the proposed GNC-enabled functions have shown a greater robustness to both poor initialization and outliers.

\section{Conclusion}
\label{sec:conclusion}

The adaptive robust kernel presented by Barron in \cite{Barron2019General} has received ample attention for its adaptability to a wider family of problems and its ability to deal with a larger set of outlier distributions.
The GNC algorithm has been effective when solving non-convex optimization problems, but it has been limited to a specific loss function in the literature.
This paper incorporates the GNC algorithm into the adaptive robust loss functions by Barron \cite{Barron2019General} and Hitchcox and Forbes \cite{Hitchcox2022Mind}.
The proposed approach demonstrates improved robustness and convergence properties, and, more importantly, eliminates the need to choose a dedicated robust loss function for a particular problem.
By combining GNC and adaptive loss function, the proposed GNC-ADAPT and GNC-AMB loss functions are able to solve nonconvex least-squares problems with a large number of outliers and poor initialization more effectively than the GNC-based counterparts \cite{Yang2020GNC}, as well as existing adaptive loss functions \cite{Barron2019General, Chebrolu2021Adaptive, Hitchcox2022Mind}.
Also, the proposed adaptive GNC loss function can give optimal solutions with high precision.
The results from point-cloud alignment, mesh registration, and pose-graph optimization problems suggest that this approach is widely applicable to any least-squares problems in state estimation and robotics.

\section*{Acknowledgment}
\noindent
The authors would like to thank Voyis Imaging Inc. for their continued feedback and support, and DECAR group members for their insightful discussions.

\printbibliography


\vspace{-9cm}
\begin{IEEEbiography}[{\includegraphics[width=1in,height=1.25in,clip,keepaspectratio]{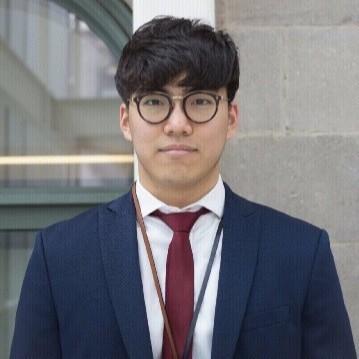}}]
    {Kyungmin Jung}
    (Graduate~Student~Member,~IEEE) received the B.Eng. and M.A.Sc. degrees in mechanical engineering in 2019 and 2021, respectively, from McGill University, Montreal, QC, Canada, where he is currently working toward the Ph.D. degree with the Department of Mechanical Engineering.
    His research interests include state estimation, computer vision, and robust optimization.
\end{IEEEbiography}
\vspace{-9cm}
\begin{IEEEbiography}[{\includegraphics[width=1in,height=1.25in,clip,keepaspectratio]{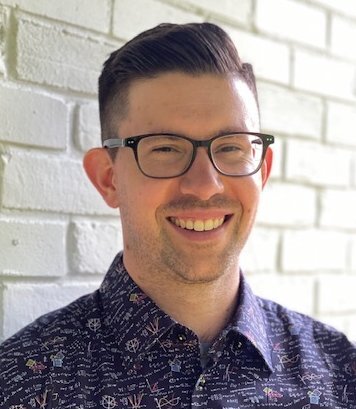}}]
    {Thomas Hitchcox}
    (Member, IEEE) received the B.Eng., M.Eng., and Ph.D. degrees in mechanical engineering from McGill University, Montreal, QC, Canada, in 2015, 2018, and 2023, respectively.
    He is currently a Research Scientist in underwater navigation and mapping at Sonardyne.
    His research interests include state estimation, computer vision, and robust algorithms for point cloud filtering and alignment.
\end{IEEEbiography}
\vspace{-9cm}
\begin{IEEEbiography}[{\includegraphics[trim={5cm 0 5.1cm 0},width=1in,height=1.25in,clip,keepaspectratio]{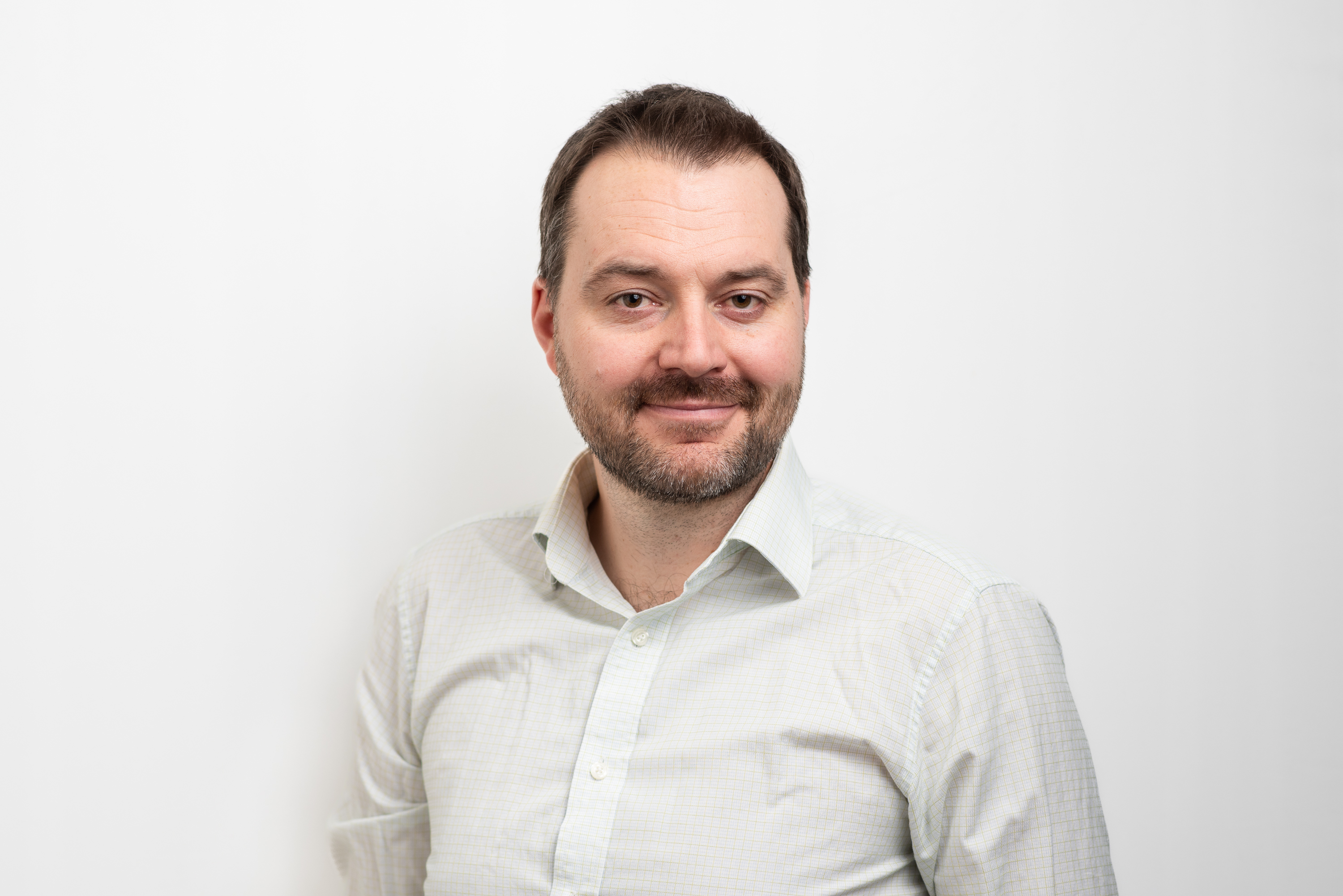}}]
    {James Richard Forbes}
    (Member, IEEE) received the B.A.Sc. degree in mechanical engineering (Honors, Co-op) from the University of Waterloo, Waterloo, ON, Canada, in 2006, and the M.A.Sc. and Ph.D. degrees in aerospace science and engineering from the University of Toronto Institute for Aerospace Studies, Toronto, ON, in 2008 and 2011, respectively.
    He is currently an Associate Professor and William Dawson Scholar with the Department of Mechanical Engineering, McGill University, Montreal, QC, Canada. His research interests include navigation, guidance, and control of robotic systems.
    Dr. Forbes is a Member of the Centre for Intelligent Machines, a Member of the Group for Research in Decision Analysis, and a Member of the Trottier Institute for Sustainability in Engineering and Design. He was the recipient of the McGill Association of Mechanical Engineers Professor of the Year Award in 2016, the Engineering Class of 1944 Outstanding Teaching Award in 2018, the Carrie M. Derick Award for Graduate Supervision and Teaching in 2020, and the Samuel and Ida Fromson Outstanding Teaching Award in Engineering in 2024. He is currently an Associate Editor for the International Journal of Robotics Research (IJRR).
\end{IEEEbiography}



\appendices

\end{document}